\documentclass[Journal]{IEEEtran}
\IEEEoverridecommandlockouts
\usepackage{enumitem}
\usepackage{titlesec}
\titlespacing*{\subsection}{0pt}{0.3\baselineskip}{0.3\baselineskip}
\usepackage{cite}
\usepackage{amsmath,amssymb,amsfonts}
\usepackage{algorithmic}
\usepackage{graphicx} \usepackage{subcaption}
\usepackage{textcomp}
\usepackage{xcolor}
\usepackage{makecell}
\usepackage{booktabs}
\usepackage{dsfont}
\usepackage{caption}
\usepackage{subcaption}
\usepackage{cite}
\usepackage{amsmath,amssymb,amsfonts}
\usepackage{algorithm,algorithmic}
\usepackage{graphicx}
\usepackage{textcomp}
\usepackage{xcolor}
\usepackage{makecell}
\usepackage{booktabs}
\usepackage{dsfont}
\usepackage{caption}
\usepackage{subcaption}
\usepackage{booktabs,siunitx}
\usepackage{mathtools}
\usepackage{comment}
\usepackage{multirow}
\def\BibTeX{{\rm B\kern-.05em{\sc i\kern-.025em b}\kern-.08em
    T\kern-.1667em\lower.7ex\hbox{E}\kern-.125emX}}
    
\ifCLASSINFOpdf
 
\else
 
\fi

\begin{document}

\title{SafeSpace \textbf{MFNet}: Precise and Efficient \textbf{MultiFeature} Drone Detection \textbf{Network}
}
\author{Misha Urooj Khan, Mahnoor Dil, 
 Muhammad Zeshan Alam, Farooq Alam Orakazi, Abdullah M. Almasoud, \\Zeeshan Kaleem,~\IEEEmembership{Senior Member,~IEEE}, Chau Yuen,~\IEEEmembership{Fellow,~IEEE}
\thanks{Misha Urooj Khan, Mahnoor Dil, Zeeshan Kaleem, Farooq Alam Orakazi are with the Department of Electrical and Computer Engineering, COMSATS University Islamabad, Wah Campus, Pakistan. (e-mail: \{mishauroojkhan, noorijazhussain, zeeshankaleem, farooqorakzai\}@gmail.com)

Muhammad Zeshan Alam is with the Department of Computer Science, Brandon University, Canada. (e-mail: alamz@brandonu.ca)

Abdullah M. Almasoud is with the Department of Electrical Engineering, College of Engineering in Al-Kharj, Prince Sattam Bin Abdulaziz University, Al-Kharj 11942, Saudi Arabia.
(e-mail: am.almasoud@psau.edu.sa)

Chau Yuen is with the School of Electrical and Electronics Engineering, Nanyang Technological University, Singapore. (e-mail: chau.yuen@ntu.edu.sg)

This work is supported by the Higher Education Commission (HEC) Pakistan under the NRPU 2021 Grant\#15687 and also by funding from Prince Sattam bin Abdulaziz University project number (PSAU/2023/R/1444).}}
%Chau Yuen,~\IEEEmembership{Fellow,~IEEE},\\ Abbas Jamalipour,~\IEEEmembership{Fellow,~IEEE}}

%\author{\IEEEauthorblockN{1\textsuperscript{st} Mahnoor Dil}
%\IEEEauthorblockA{\textit{Comsats University Islamabad} 
%\\ Wah Campus, Pakistan
%\\noorijazhussain@gmail.com}
%\and
%\IEEEauthorblockN{2\textsuperscript{nd} Misha Urooj Khan}
%\IEEEauthorblockA{\textit{Comsats University Islamabad} 
%\\ Wah Campus, Pakistan
%\\ mishauroojkhan@gmail.com}
%\and
%\IEEEauthorblockN{3\textsuperscript{rd} Farooq Alam Orakazi}
%\IEEEauthorblockA{\textit{Comsats University Islamabad}
%\\ Wah Campus, Pakistan
%\\email address or ORCID}
%\and
%\IEEEauthorblockN{4\textsuperscript{th} Zeeshan Kaleem}
%\IEEEauthorblockA{\textit{Comsats University Islamabad} 
%\\ Wah Campus, Pakistan
%\\zeeshankaleem@ciitwah.edu.pk}
%}
\maketitle

\begin{abstract}
The increasing prevalence of unmanned aerial vehicles (UAVs), commonly known as drones, has generated a demand for reliable detection systems. The inappropriate use of drones presents potential security and privacy hazards, particularly concerning sensitive facilities. Consequently, a critical necessity revolves around the development of a proficient system with the capability to precisely identify UAVs and other flying objects even in challenging scenarios. Although advancements have been made in deep learning models, obstacles such as computational intricacies, precision limitations, and scalability issues persist. 
To overcome those obstacles, we proposed the concept of \textit{MultiFeatureNet} (MFNet), a solution that enhances feature representation by capturing the most concentrated feature maps. Additionally, we present \textit{MultiFeatureNet-Feature Attention} (MFNet-FA), a technique that adaptively weights different channels of the input feature maps. To meet the requirements of multi-scale detection, we presented the versions of MFNet and MFNet-FA, namely the small (S), medium (M), and large (L). The outcomes reveal notable performance enhancements. For optimal bird detection, MFNet-M (Ablation study 2) achieves an impressive precision of 99.8\%, while for UAV detection, MFNet-L (Ablation study 2) achieves a precision score of 97.2\%. Among the options, MFNet-FA-S (Ablation study 3) emerges as the most resource-efficient alternative, considering its small feature map size, computational demands (GFLOPs), and operational efficiency (in frame per second). This makes it particularly suitable for deployment on hardware with limited capabilities. Additionally, MFNet-FA-S (Ablation study 3) stands out for its swift real-time inference and multiple-object detection due to the incorporation of the FA module. The proposed MFNet-L with the focus module (Ablation study 2) demonstrates the most remarkable classification outcomes, boasting an average precision of 98.4\%, average recall of 96.6\%, average mean average precision (mAP) of 98.3\%, and average intersection over union (IoU) of 72.8\%. 
To encourage reproducible research, the dataset, and code for MFNet are freely available as an open-source project: github.com/ZeeshanKaleem/MultiFeatureNet.
\end{abstract}

\section{Introduction}
\begin{figure}[h]
  \begin{center}
  \includegraphics[width=2.5 in]{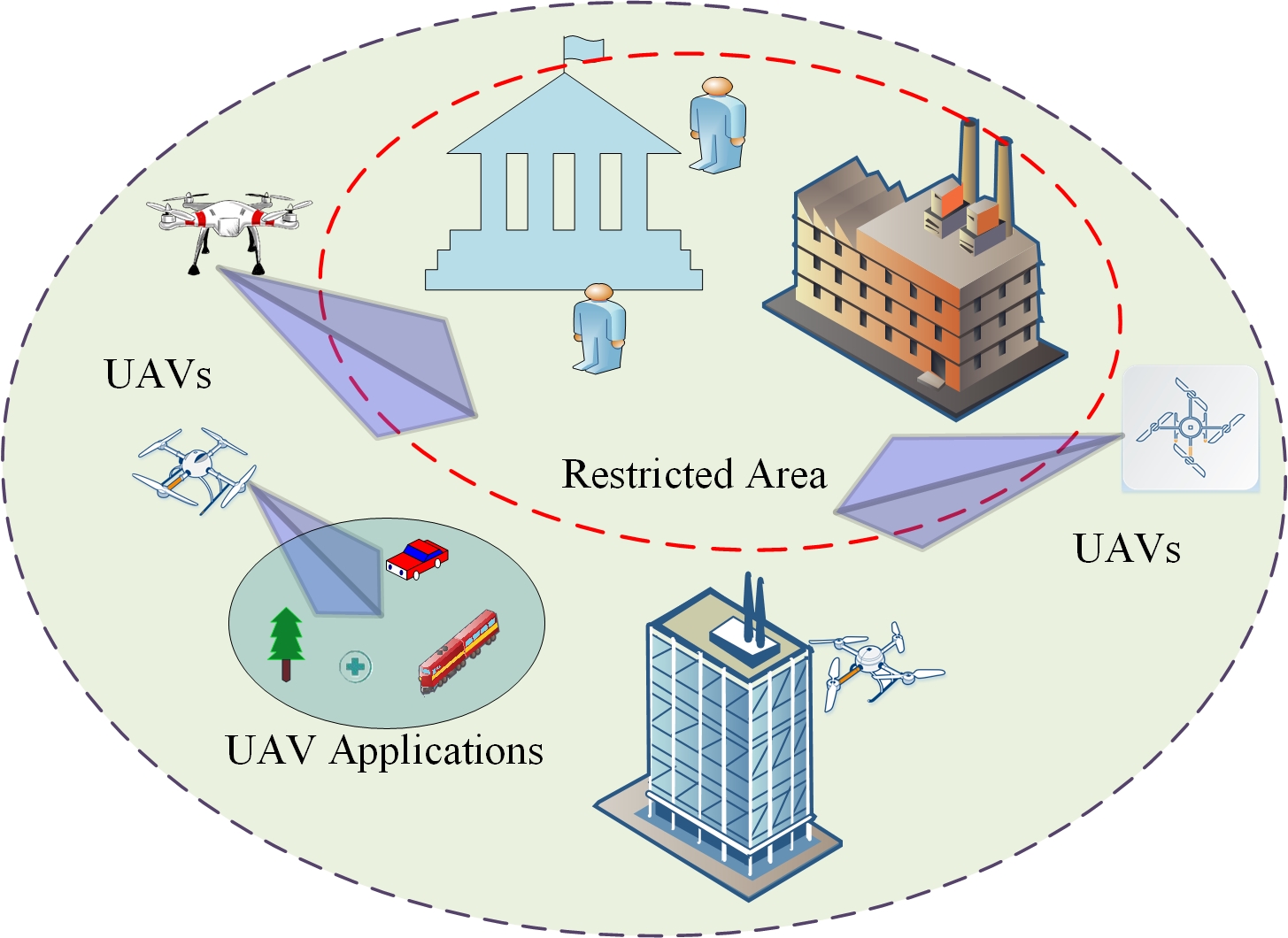}
  \caption{UAV's applications and its implications in security-sensitive areas. UAVs have applications in numerous fields ranging from performing complex tasks for the military to delivering food in homes \cite{i2}.}
  \label{fin}
   \end{center} 
\end{figure}

The market of unmanned aerial vehicles (UAVs), commonly known as drones, was valued at USD 10.72 billion in 2019, and it is expected to grow to USD 25.13 billion by 2027 due to the increasing demand for automation and advancements in technology \cite{i1}. Due to their low cost and ease of operation, UAVs are being utilized for various tasks such as surveillance, healthcare, animal tracking, and disaster response, as depicted in Fig. \ref{fin}. However, UAVs can threaten national security by violating security protocols and entering sensitive areas, allowing unauthorized organizations or individuals to transport hazardous materials \cite{i2}. This misuse of drones jeopardizes public safety and security and increases the risk of aerial accidents in case of mid-air collisions at low speeds. Therefore, there is a critical need for automated drone detection technology that can prevent unnecessary drone interventions, detect and deactivate unknown drones quickly, and enhance public safety and security. 

Kaleem \textit{et al.} have reported that although drone neutralization approaches have improved their effectiveness, they rely on expensive and specialized equipment \cite{i8}. Hence, there is a need for cost-effective hardware-based drone detection systems that offer high performance. Typically, two approaches are employed for UAV detection: Ground-to-Air Detection (GAD), which deploys cameras on the ground to identify UAVs in flight, and Air-to-Air Detection (AAD), in which a flying UAV uses onboard cameras to detect other UAVs in flight. In most GAD activities, ground cameras are stationary or move passively, and the target UAV is shot against a bright or cloudy sky background. In contrast, a flying UAV in AAD deployment captures the target UAV from a top or side view, where the target UAV's background features complex scenes like urban or natural settings. The onboard camera is flying dynamically, leading to significant changes in the target UAV's appearance, such as its form, scale, and color \cite{i3}. Consequently, this method yields less performance compared to GAD.

According to Lykou \textit{et al.} \cite{i4}, commercial UAV detection systems predominantly rely on visual sensors (40\%), followed by radar-based systems (28\%), and radio frequency (RF) sensors (26\%). Acoustic sensors constitute a minority (6\%) of such systems. Various sensors have been utilized for detecting and classifying drones, including radar, which employs active and passive sensing over multiple frequency bands \cite{i5}; visible spectrum cameras \cite{i3}; thermal or infrared (IR) cameras \cite{i6}; LIDAR (light detection and ranging) \cite{i7}; microphones or acoustic sensors for detecting acoustic vibrations \cite{i2}; and RF monitoring sensors that detect radio signals \cite{yuen18,i9,lr3}. Detection methods for UAVs using acoustics and RF sensing have limitations such as higher costs or low range and accuracy of detection, as noted by previous studies \cite{i5}. On the other hand, camera-based detection methods that use visible images have gained popularity due to their high resolution, which allows for efficient classification and object recognition. However, challenges such as lighting changes, occluded sections, and cluttered backgrounds make camera-based detection methods less effective, necessitating the exploration of more efficient methods. The current literature lacks significant research on using thermal and IR cameras for UAV detection in challenging weather conditions \cite{i6}. Furthermore, sensor fusion \cite{i7},\cite{xie2020adaptive} is a promising research area that could enhance UAV detection accuracy. 

 \subsection{Challenges and limitations}
With advanced hardware possessing superior accelerated technologies, accurate and robust UAV detection is achievable. Convolutions neural network (CNN), employs visual and acoustic information to classify UAVs \cite{lr3,lr4} and has demonstrated superior feature extraction capabilities compared to traditional object recognition algorithms. More complex object detection models, such as YOLO (you only look once) \cite{lr4}, exhibit exceptional object recognition precision and speed compared to others. They are faster than region-based techniques due to their simplified architectural design. Despite these advancements, the proposed research methodologies in the existing literature have encountered numerous challenges during drone detection and classification.
 \begin{itemize}
   \item \textbf{UAVs vs. birds classification:} UAVs are physically similar to birds, which generates the false alarm for birds during the UAVs identification stage. 
    \item \textbf{Crowded backgrounds:} The inability to accurately segregate UAVs from the background makes UAVs detection difficult when present in a dense /cluttered background having clouds, flames, mist, sun, and smoke in the sky.
     \item \textbf{Different-size UAVs:} It's quite exigent to train a deep learning model which is sensitive to different-size UAVs \cite{i7}.
 \end{itemize}

 \vspace{-2mm}
 \subsection{Contributions}
In this paper, we propose a precise and efficient multi-feature and multi-scale UAV detection network, i.e., SafeSpace \textit{MultiFeatureNet} (MFNet). We address the pre-discussed gaps by developing an open dataset and proposing a real-time detection model which can excellently classify birds vs. UAVs with speedy inference time via attention given to the most important feature maps. The key contributions are summarized here:

\begin{itemize}  
    \item A carefully chosen dataset has been created for addressing the drones versus birds issue. This dataset encompasses a range of environmental settings, multi-sized objects, varying drone dimensions, and multiple object types.
    \item Introduced the \textbf{Feature attention module}, which computes attention weights by leveraging learned feature correlations. This empowers MFNet to dynamically highlight crucial channels while diminishing less insightful ones thereby enhancing the overall feature representation and improving the model's performance.
    \item Integration of the \textbf{Focus module} to capture the attributes of smaller targets enhances detection accuracy. It also prevents the potential feature deterioration resulting from restricted pixel representation in the original images. Simultaneously, the module maintains detection precision for larger targets present in the images. 
    \item Incorporation of advanced techniques like dynamic batch size adjustment (DBSA) and automated machine learning (AutoML), Scaled weight decay factor (SWDF) for optimal training with the extraction of most focused feature maps. 
    \item Three distinct versions of MFNet/MFNet-FA, namely small (S), medium (M), and large (L), have been examined across three separate ablation studies. This leads up to nine proposed models under evaluation.
   \end{itemize}

The rest of the paper is structured as follows: Section II summarized the literature review by critically analyzing the existing schemes and their shortcomings. In section III, the proposed MFNet/MFNet-FA architecture is presented. However, in section IV, the dataset, implementation, and key findings of the proposal have been evaluated. Furthermore, Section V discusses the performance evaluation of the considered ablation studies. Finally, in Section VI the conclusion of the paper is presented.
\vspace{-3mm}
\section{Literature Review}
Li \textit{et al.} in \cite{i1} proposed software-defined radio (SDR) for detecting and classifying jamming attacks on UAVs. They used conventional and deep learning algorithms with 35\% and 0.03\% false alarm rates, respectively, for spectrogram-based classification. Mel-frequency and linear predictive cepstral coefficients with support vector machines (SVM) achieved 96.7\% accuracy \cite{i2}. Zheng \textit{et al.} \cite{i3} used monocular cameras and introduced the Det-Fly dataset for aerial anomaly detection (AAD) with the highest accuracy of 82.4\% for vision-based swarm and malicious UAV detection. The authors in \cite{i5} used acoustic, image/video, and wireless RF signals for UAV classification and detection with a hybrid synthetic framework. Gerwen \textit{et al.} \cite{i7} used a flexible sensor fusion platform to target multiple sensors' influence on 3D indoor location accuracy.
Alsanad \textit{et al.} improved YOLOv3 for better drone detection accuracy by incorporating dense connecting modules and multiple-scale detection. They achieved a high accuracy of 95.60\%, 0.36 mAP, and 60 FPS \cite{lr4}. LIDAR-assisted UAV detection was introduced in \cite{lr1} to detect and track drones of different sizes. The authors in \cite{lr2} used SqueezeNet to extract multiple features from an RF dataset for a range of signal-to-noise ratio (SNR) [5-30]dB, showing a significant improvement compared to conventional classifiers like KNN and SVM.
Wisniewski \textit{et al.} \cite{lr5} utilized CNN for real-time anti-UAV demonstration videos by varying parameters like model orientation, backdrop graphics, and textures via domain randomization to provide a pixel-level mask of the drone's position, saving time spent on individual drone labeling. These approaches demonstrate effective UAV detection and classification systems for various scenarios.
{Dai\textit{et al.} trained a pre-trained YOLOv5s model to calculate the in-front drone's relative position, demonstrating its sensor capabilities \cite{lr7}. In \cite{lr8}, authors adopted a result-level fusion-based 2D-CNN for binary classification with audio signals. The proposed DIAT-RadSATNet contains modules from SqueezeNet and MobileNet for multi-class classification \cite{lr9}. Elsayed \textit{et al.} presented a visual drone detection method using videos with uniform backgrounds. They proposed a CNN classifier's background removal algorithm in the detection phase and used the tracking phase for missed detection tracks \cite{lr10}. GLF-Net, a global-local feature-enhanced network with a multiscale feature fusion module, was proposed in \cite{lr11}. The authors achieved a detection accuracy of 86.52\% mAP on the rotated object detection UAV (RO-UAV) dataset. Here, Ye \textit{et al.} introduced CT-Net, a convolution-transformer network that utilized an attention-enhanced transformer block and a feature-enhanced multi-head self-attention to perform low-altitude object detection. The model achieved impressive results, with an mAP score of 0.966 on small objects, outperforming the baseline YOLOv5 \cite{lr12}. Training object detection algorithms, especially for UAV detection, face the challenge of hierarchical feature extraction to obtain multilevel representations from pixels to high-level semantic features. Multilevel nonlinear mappings result in tangled hidden factors, reducing the model's ability to express real-time and unseen images. In addition, some algorithms struggle with high-dimensional data that has a multi-class problem.

However, upon reviewing the literature, we found limitations in existing datasets used for training, which only contained one type of drone image and ignored challenging weather conditions or complex environments. Furthermore, details on the number of samples per class were often missing, and data augmentation techniques to balance imbalanced classes were ignored. Finally, the training time and evaluation metrics, such as mAP, and IoU, were often neglected, which are crucial for determining the necessary computational resources. These limitations and challenges have motivated us to propose MFNet, which aims to improve precision, efficiently target multiscale UAVs, and provide balanced results in challenging weather conditions.
\begin{figure*}[ht]
  \begin{center}
  \includegraphics[width=7in]{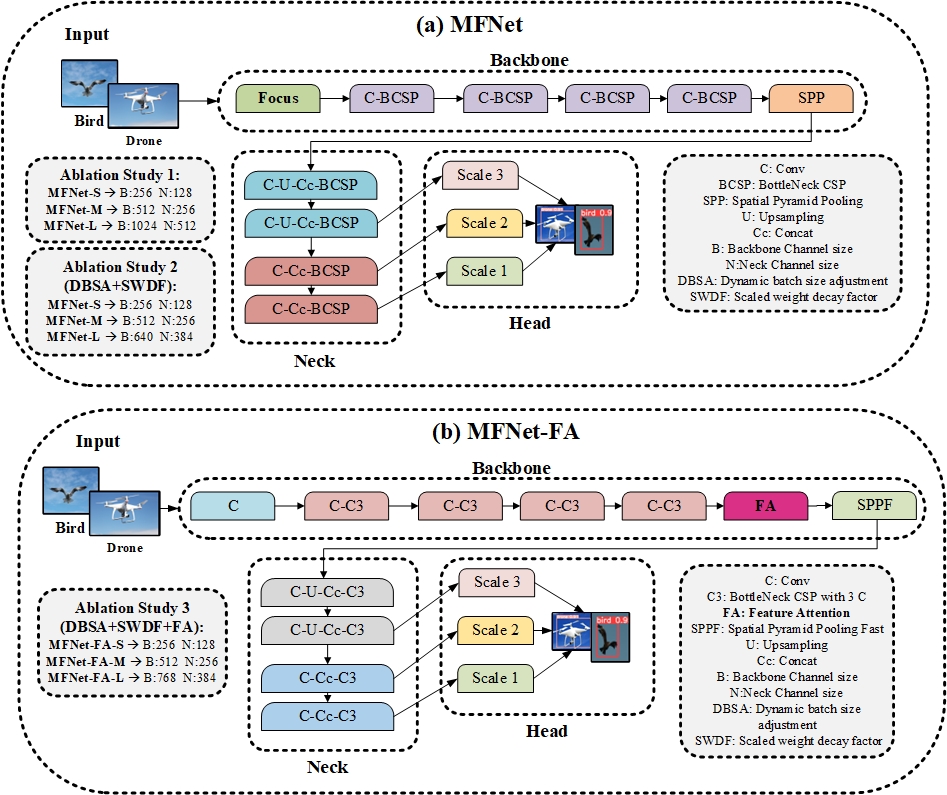}
  \caption{Proposed \textbf{SafeSpace MFNet}. (a) MFNet: Takes fixed input size of $320\times320$; extracts \textbf{focused feature maps} by using \textbf{focus module} in backbone block. Focused feature maps are then fused in the head block, and the final prediction is made using the head layer, which o draws bounding boxes around the identified objects. (b) MFNet-FA: novel \textbf{Feature attention module} is added in the backbone block that computes attention weights through feature associations, allowing the MFNet-FA to dynamically highlight the most significant/attentive feature maps and channels.}
  \label{f2}
  \end{center}   
\end{figure*}

\section{Proposed MFNet/MFNet-FA Architecture}
\subsection{MFNet/MFNet-FA blocks description}
The proposed MFNet/MFNet-FA has four main blocks: input, backbone, neck, and head, as shown in Fig. \ref{f2}. The \textit{input block} transfers spatial information to the channel dimension on the input images for faster inference with no mAP penalty. The \textit{backbone} retrieves attentive or most important feature maps of various sizes from the input image by using cross-stage partial network (CSP) \cite{f1} and spatial pyramid pooling (SPP) \cite{f2}. In MFNet, we integrate \textbf{Focus} module to focus on the most significant subsets of spatial information with reduced spatial dimensions to focus on unseen features. Unseen features in the context of computer vision refer to subtle details or characteristics that are inherently challenging to detect, especially in very small objects. This helps capture different scales and patterns in the input image while reducing computational complexity.
BottleneckCSP block in the backbone reduces the computation/inference time, and SPP improves detection accuracy for three-scaled feature maps. 

However, in the proposed MFNet-FA, we integrated BottleneckCSP with three convolution layers (3C) and spatial pyramid pooling fast (SPPF) with a \textbf{feature attention} module. This combination enables the model to calculate attention-based feature maps with down-sampled parameters, which makes it sensitive to fin-grain details compared to MFNet. For the \textit{neck block}, \textit{feature pyramid network (FPN)} \cite{f3} is used, which extracts semantic qualities from top to lower hierarchy, whereas \textit{path aggregation network (PAN)} \cite{f4} extract localization features from lower to top order. These two structures collaborate to strengthen the features acquiring capability from multiple network levels by fusion, helping in increasing detection capabilities even more. The last block, named \textit{head}, performs the final detection. MFNet/MFNet-FA facilitates detection across three different scales achieved through downsampling the input image dimensions by factors of 32, 16, and 8, respectively. The initial detection occurs using the feature map from the 17th layer, followed by the second detection at the 20th layer, and ultimately concluding with the third and final detection utilizing the feature map from the 23rd layer.

\subsection{Anchor boxes}
An anchor box is a predetermined collection of bounding boxes with a specific height and breadth. These boxes are chosen based on the object sizes in training datasets to capture various object classes' scale and aspect ratio. The anchor boxes for MFNet/MFNet-FA are $8\times8$ (P3), $16\times16$ (P4), and $32\times32$ (P5) for the identification of smaller, medium, and large objects. 

\subsection{Loss functions and target prediction}
To optimize MFNet/MFNet-FA}, we utilize adaptive moment estimation (Adam) optimizer \cite{f6} and loss functions introduced by YOLOv5 \cite{lr12} named as \textit{class loss $(L_{cls})$}, \textit{objectness loss $(L_{obj})$}, and \textit{localization loss $(L_{loc})$}. Adam computes the decaying averages of past $mt$ and the past squared gradients of $vt$.
\begin{equation}
m_1= \gamma_1\times m_{1_{(t-1)}} + (1-\gamma_1)\times \nabla_t,
\end{equation}
\begin{equation}
m_2= \gamma_2\times m_{2_{(t-2)}} + (1-\gamma_2)\times \nabla_t^2,
\end{equation}
where $m_1$ is the first moment (the
mean), $m_2$ is the second moment (the uncentered variance) of the gradients, $\gamma_1$ is the exponential decay rate for the $m_1$, $\gamma_2$ is the exponential decay rate for $m_2$, and $\nabla_t$ is the gradient of the current mini-batch. These estimates update the parameter $\theta$ using the following equation as 
\begin{equation}
\theta_t=  \theta_{(t-1)} - \frac{\eta}{\sqrt{\hat{m_2}+\in}}+\hat{m_1}.
\end{equation}
$\eta$ represents the learning rate, and the feature map of all MFNet/MFNet-FA models is represented by $g(f(I_x,y,t))$ having $Z$x$Z$ grids in each cell, have $B$ bounding boxes where prediction losses are applied. The $L_{obj}$ can only consider during the presence of an object in a particular cell $C$.
\begin{equation}
L_{obj}= \sum_{l=0}^{Z^2}\sum_{m=0}^{B}1_{obj}^{lm}(C_l-\hat{C_l})+\lambda_{n}\sum_{l=0}^{Z^2}\sum_{m=0}^{B}1_{obj}^{lm}(C_l-\hat{C_l}),
\end{equation}
where $1_{obj}^{lm}$ has a value of 1 if the $m$-th bounding box in cell $l$ contains the object. $\lambda_{n}$ is loss coefficient. $L_{cls}$ is computed using a square of the error between predicted conditional class probability $\hat{\mathds{P}}_l(c)$ and ground-truth $\mathds{P}_l(c)$ for cell $l$.
\begin{equation}
L_{cls}= \sum_{l=0}^{Z^2}1_{obj}^{lm}\sum_{c \in C}(\mathds{P}_l(c)-\hat{\mathds{P}}_l(c)),
\end{equation}
 Here, $L_{loc}$ returns the difference between the predicted and true bounding boxes, and is computed as
\begin{equation}
\begin{aligned}
L_{loc}= \sum_{l=0}^{Z^2}\sum_{m=0}^{B}1_{obj}^{lm}\left[(x_l-\hat{x}_l)^2+(y_l-\hat{y}_l)^2\right] +\\ \lambda_{cd}\sum_{l=0}^{Z^2}\sum_{m=0}^{B}1_{obj}^{lm}\left[(\sqrt{w_l}-\sqrt{\hat{w}_l})^2+(\sqrt{h_l}-\sqrt{\hat{h}_l})^2\right].
\end{aligned}
\end{equation}
$(\hat{x}_l, \hat{y}_l)$ and $(x_l, y_l)$ represent the top-left corner coordinates of the predicted and ground truth bounding box, respectively. Whereas $(\hat{w}_l, \hat{h}_l)$ and $({w}_l, {h}_l)$ are the width and height of predicted and ground truth bounding box, respectively. Therefore, the total loss for the MFNet/MFNet-FA model is
\begin{equation}
L_{t}= \lambda_1 \times L_{cls}+ \lambda_2 \times L_{obj}+ \lambda_3 \times L_{loc}.
\end{equation}
These are the MFNet/MFNet-FA losses: $L_{obj}$ is the confidence of object existence calculated using binary cross-entropy loss. Similarly, $L_{loc}$ is the bounding box regression loss calculated using mean squared error, and $L_{cls}$ is the classifying loss calculated using cross-entropy. Here, $\lambda_{1}$, $\lambda_{2}$ and $\lambda_{3}$ are loss coefficients. 

MFNet/MFNet-FA calculates the bounding box's target coordinates and target frame size with a specific grid size. The network predicts 4 coordinates for each bounding box, $t_x, t_y, t_w, t_h$. If the cell is offset from the top left corner of the image by $(cx, cy)$ and the bounding box prior has width and height $pw $, $ph$, and $\sigma$ is the sigmoid activation function, then the predictions correspond to:
\begin{equation}
b_x = \left(2 \cdot \sigma(t_x)-0.5\right)+ c_x.
\end{equation}
\begin{equation}
b_y = \left(2 \cdot \sigma(t_y)-0.5\right)+ c_y.
\end{equation}
\begin{equation}
b_w = p_w \cdot \left(\sigma(t_w)\right)^{2}.
\end{equation}
\begin{equation}
b_h = p_h \cdot \left(\sigma(t_h)\right)^{2}.
\end{equation}
\vspace{-4mm}
\subsection{Focus module}
The Focus module, as introduced in \cite{10189370}, serves as the central component in MFNet, thoughtfully crafted to elevate the representation and extraction of focused feature maps. Its primary objective is to enhance the detection of previously overlooked very fine-grained details, often referred to as unseen features. The input tensor is partitioned into four segments via spatial segmentation denoted as $x_{\text{in}} \in \mathbb{R}^{b \times c_1 \times w \times h}$, each encapsulating specific spatial nuances. These segmented tensors are then combined along the channel dimension, resulting in a concatenated tensor denoted as $x_{\text{cat}} \in \mathbb{R}^{b \times 4c_1 \times \frac{w}{2} \times \frac{h}{2}}$.

The Sigmoid Linear Unit (SiLU) is employed to modulate interactions between tensor elements, thereby amplifying its nonlinear expressive capabilities. This activated tensor then undergoes convolutional processing. Here, a convolutional layer, defined by parameters like kernel size (k), stride (s), padding (p), and groups (g), operates on the concatenated tensor $x_{\text{cat}}$. This transformative convolutional operation is once again catalyzed by the SiLU, introducing nonlinearity and enhancing the extraction of the most pertinent patterns within the concatenated tensor. Ultimately, this process generates focused feature maps characterized by enriched spatial representations and the finest details.
\begin{comment}
\begin{algorithm}[!h]
\caption{\textcolor{blue}{Focus Module Pseudocode}}
\label{alg1}
\begin{algorithmic}[1]
\REQUIRE $c_1$, $c_2$, $k$, $s$, $p$, $g$, $act$
\STATE $x_{\text{in}} \in \mathbb{R}^{b \times c_1 \times w \times h}$ \Comment{(input feature maps)}
\STATE $x_1 = x_{\text{in}}[..., ::2, ::2]$
\STATE $x_2 = x_{\text{in}}[..., 1::2, ::2]$
\STATE $x_3 = x_{\text{in}}[..., ::2, 1::2]$
\STATE $x_4 = x_{\text{in}}[..., 1::2, 1::2]$
\STATE $x_{\text{cat}} = \text{concat}(x_1, x_2, x_3, x_4, \text{dim}=1)$
\STATE $x_{\text{cat}} \in \mathbb{R}^{b \times 4c_1 \times \frac{w}{2} \times \frac{h}{2}}$
\STATE \text{act} = \text{SiLU()}
\STATE $\text{output} = \text{Conv}(x_{\text{cat}}, c_2, k, s, p, g, act)$ 
\STATE \textbf{Output:} Focused feature maps
\end{algorithmic}
\end{algorithm}
\end{comment}

\subsection{Feature attention module}
The Feature Attention (FA) module enhances the feature representation of MFNet-FA by adaptively weighting different channels of the input feature maps. Mathematically, given an input tensor with dimensions (batch size $b$, number of channels $c$, height $h$, width $w$), the FA module starts by applying global average pooling to reduce the spatial dimensions to 1$\times$1, resulting in a tensor $y$ of size ($b, c, 1, 1$). Next, the module employs two linear transformations to learn feature relationships. It applies a linear layer with $\frac{c}{ratio}$ output units, where $ratio$ considered here equals 16, followed by a ReLU (rectified linear unit) activation function. Afterward, an additional linear layer is added with $c$ output units. It uses a sigmoid activation function to normalize the importance ratings. The sigmoid-activated attention weights are then applied to each channel, reshaped to a 1$\times$1 tensor, and multiplied element-wise with the original input feature maps $x$. This element-wise multiplication emphasizes the most informative features and suppresses redundant/less-important ones, effectively focusing the MFNet-FA's attention on critical details. 
%Below is the pseudo-code of the proposed FA module \ref{alg1},%
$c_1$ represents the number of input channels in the input feature maps, $c_2$ is the number of output channels in the feature maps after applying the FA module, and $ratio$  is the reduction ratio in the linear transformations.
\begin{comment}
\begin{algorithm}[!h]
\caption{\textcolor{blue}{Proposed Feature Attention Module (FA) pseudocode}}
\label{alg1}
\begin{algorithmic}[1]
\REQUIRE $c_1$, $c_2$, $ratio$, $c$, $b$, $h$, $w$
\STATE $x \in \mathbb{R}^{b \times c \times h \times w}$ (input feature maps)
\STATE $c_{\text{out}} = \lfloor \frac{c_1}{ratio} \rfloor$
\STATE $y = \text{avgpool}(x) \in \mathbb{R}^{b \times c \times 1 \times 1}$ 
\STATE $y = \text{l1}(y) \in \mathbb{R}^{b \times c_{\text{out}}}$ 
\STATE $y = \text{ReLU}(y) \in \mathbb{R}^{b \times c_{\text{out}}}$ 
\STATE $y = \text{l2}(y) \in \mathbb{R}^{b \times c_1}$
\STATE $y = \text{Sigmoid}(y) \in \mathbb{R}^{b \times c_1}$ 
\STATE $y = y \in \mathbb{R}^{b \times c_1 \times 1 \times 1}$ 
\STATE $\text{output} = x \cdot y \in \mathbb{R}^{b \times c \times h \times w}$
\STATE \textbf{Output:} Attention feature maps
\end{algorithmic}
\end{algorithm}
\end{comment}

\subsection{MFNet/MFNet-FA working procedure}
MFNet/MFNet-FA processes the entire image using a single neural network,  divides it into grids and predicts bounding boxes with probabilities for each grid cell. The predicted probability weights the bounding boxes as it provides predictions after only one forward propagation passes through the neural network. Lastly, the max suppression algorithm ensures that the MFNet identifies each object once. The network topology in Fig. \ref{f2} is the same for all MFNet/MFNet-FA models. We proposed MFNet/MFNet-FA models to compute the optimum size of feature maps that provides the best and fast detection results for flying birds and UAVs in challenging weather conditions. The anchor sizes, model depth multiple (0.33), and layer channel multiple (0.50) are the same for all proposed models.

\section{Dataset and Implementation}
UAVs are challenging to detect because of their closeness to birds in terms of radar cross-section (RCS), moderate velocities, and low flying altitudes. Generally, birds are misidentified as UAV targets by the drone surveillance system resulting in an unnecessarily high incidence of false reports and lowering the efficacy of the surveillance technique. To overcome this, we consider the birds vs. UAV detection problem to improve the system's precision and mAP compared to the existing schemes. 
\subsection{Datset and Hardware Resources}
To train the proposed MFNet/MFNet-FA architecture, \textbf{5105 images} of UAVs and birds from the publically available open-source datasets are collected. The collective dataset has various types of UAVs like multi-rotor (tri, quad, Hexa, and octa-copter), single rotor, fixed wings, and various types of birds. The images also contain different-sized (small, medium, and large) birds and UAVs. We target the detection of bird and UAVs problem with the target that MFNet/MFNet-FA will be sensitive to multi-size and multi-types of birds and UAVs. Among 5105 images, UAVs have 2605 images, and birds have 2500. Of these images, (850, 833) are small-sized, (855, 833) medium-sized, and (900, 834) large-sized (birds, drone) images. To fulfill the muti-type UAV and birds criteria, these small-, medium-, and large-sized images contain different drone models and different types of birds (white stork, crane, rüppell's vulture, eagle, bar-tailed godwit, and common blackbird). 

\begin{figure}[ht]
  \begin{center}
  \includegraphics[width=3.5in]{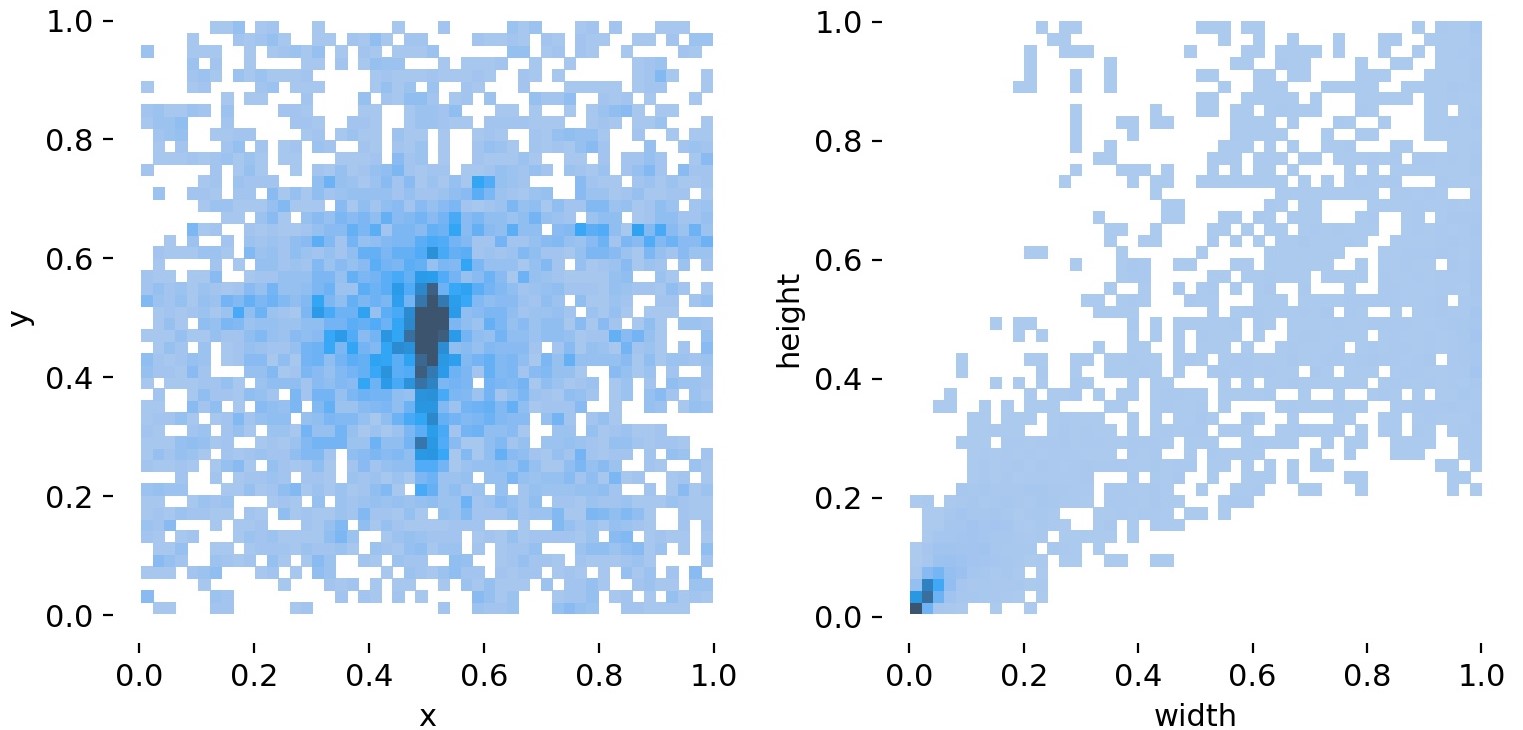}
  \caption{Dataset Distribution.}
  \label{f3}
  \end{center}   
\end{figure}

The dataset contains 8 backgrounds clear sky, cloudy, sunny, fog, rainy, water, mountains, and forest. We have approximately 325 UAV images per background (e.g., cloudy background) and 312 bird images per background. For dataset pre-processing, contrast enhancement is applied to get a fixed range of intensity values to reduce the intensity spread, training time, and exponential increase in computational resources. All the datasets are annotated and imported in the \textit{Roboflow- YOLO Darknet TXT format} for training the models. We split the total 5105 pre-processed images into 4340 training images \textbf{(85\%)}, 510 validation images \textbf{(10\%)}, and 255 test images \textbf{(5\%)}. We plot the distribution of the drone sizes and positions in the dataset in Fig. \ref{f3}. All experiments are separately run on the \textit{Google Colab} environment with an NVIDIA Tesla T4 GPU and 12GB RAM. The hyper-parameters for all models are set at the same values for a fair comparison with the state-of-the-art as given in Table \ref{t3}.
\vspace{-4mm}
\subsection{Optimal batch size selection}
Determining the perfect batch size for a model can be challenging and lead to memory crashes. However, our solution incorporates Gradient Accumulation (GA) into MFNet/MFNet-FA, which dynamically adjusts the batch size. GA can accumulate the gradients of multiple batches before updating the model weights, enabling larger batch sizes that can enhance the model's convergence rate without exceeding the GPU memory capacity. DBSA conducts a series of tests with increasing batch sizes to determine the optimal batch size for a specific GPU, measuring the corresponding training time and memory usage. This technique automatically selects the optimal batch size for MFNet based on the trade-off between training time and memory usage for conducting ablation studies 2 \& 3. With this approach, the GPU operates efficiently without running out of memory or disrupting the training process.

\subsection{Scaled weight decay factor (SWDF)}
Maintaining effective regularization control is essential for dynamically adjusting weight decay. Scaled weight decay assumes a pivotal role in accomplishing this goal. By appropriately regulating various layers and parameters according to their magnitudes, MFNet/MFNet-FA models exhibit enhanced optimization and better generalization performance. This parameter is calculated across all versions of ablation studies 2 and 3 to attain optimal outcomes.

\subsection{Optimal image size selection}
Leveraging AutoML through a heuristic search algorithm has significantly elevated the performance of MFNet/MFNet-FA. This is achieved by dynamically adjusting the input image size throughout the training process. The algorithm continuously refines the image size in response to the model's performance, enabling MFNet/MFNet-FA to seamlessly accommodate diverse image sizes and attain peak performance. Following a comprehensive examination, we have identified that an optimal image size for our dataset is 320 $\times$ 320, a dimension we have employed in ablation studies 2 and 3.

\subsection{Optimizer, learning rate, and weight decay}
We initialized the learning rate (lr0) to 0.01 for both the SGD and Adam optimizers. The momentum is adjusted to 0.937, aiming to expedite learning in directions with low curvature while maintaining stability in high-curvature directions. To counteract overfitting by penalizing excessive weights, a weight decay of 0.0005 is employed. This choice contributes to optimizer convergence by encouraging lower weights, enhancing training efficiency, and reducing convergence time. Additionally, we designated a warm-up epoch of 3.0, accompanied by initial warm-up momentum and bias values of 0.8 and 0.1, respectively. These parameter values remained consistent across all ablation studies and YOLOv5s.

\section{Performance Evaluation of Ablation Studies}
The ablation findings and detailed comparison of MFNet/MFNet-FA vs. YOLOv5 obtained after extensive experimentation are presented in Table \ref{t3}. We evaluate the proposed architecture detection performance by using the evaluation metrics; precision ($P$), recall ($R$), mean average precision ($mAP$), and intersection over Union ($IoU$), respectively. Mathematically;
\begin{equation}
    P=\frac{TP}{TP+FP}
\end{equation}
\begin{equation}
    R=\frac{TP}{TP+FN}
\end{equation}
\begin{equation}
    mAP=\frac{1}{N}\sum_{c=1}^{N}{\left(\frac{TP}{TP+FP}\right)}
\end{equation}
\begin{equation}
    IoU=\frac{\text{GroundTruth} \cap \text{DetectedBox}}{\text{GroundTruth} \cup \text{DetectedBox}}
\end{equation}

\begin{table}[ht]
\caption{Evaluation metrics and Ablation results of the trained models.}
\centering
\begin{tabular}{p{2cm}p{1cm}p{1cm}p{1.5cm}p{1cm}}
\hline
   \multicolumn{5}{c}{\textbf{(a) Ablation study 1}}
   \\ \hline 
    \textbf{Class} & \textbf{Precision (\%)} & \textbf{Recall (\%)} & \textbf{mAP (\%)}& \textbf{IoU (\%)} 
\\ \hline 
    \multicolumn{5}{c}{\textbf{MFNet-S}}
    \\ \hline
     \textbf{Bird} &86.7 &83.1 &86.3 &37.4
    \\ \hline
    \textbf{UAV} &95.2 &90 &95.4 &60.8 
    \\ \hline
    \textbf{Average} &91  &86.6 &90.8 &49.1  
    \\ \hline
    \multicolumn{5}{c}{\textbf{MFNet-M}}
    \\ \hline
    \textbf{Bird} &87.7 &86.4 & 87.1 &39.4
    \\ \hline
    \textbf{UAV} &96.8 & 90.4 &95.9 &62.7
    \\ \hline
    \textbf{Average} &92.3 &88.4 &91.5  &51.1
    \\ \hline
    \multicolumn{5}{c}{\textbf{MFNet-L}}
    \\ \hline
    \textbf{Bird} &87.4 &82.1 &86.1 &36.8 
    \\ \hline
    \textbf{UAV} &95.8 &88.2 &95.6 &61.1
    \\ \hline
    \textbf{Average} &91.6  &85.2   &90.9   &49  
   \\ \hline  
   \multicolumn{5}{c}{\textbf{(b) Ablation study 2}}
   \\ \hline 
    \textbf{Class} & \textbf{Precision (\%)} & \textbf{Recall (\%)} & \textbf{mAP (\%)}& \textbf{IoU (\%)} 
    \\ \hline
    \multicolumn{5}{c}{\textbf{MFNet-S}}
    \\ \hline
     \textbf{Bird} &98.9  &93.8 &98.3  &68.3
    \\ \hline
    \textbf{UAV} &95.4 &91.1  &94.4  &63.3
    \\ \hline
    \textbf{Average} &97.1 &92.4 &96.3 &65.8
    \\ \hline
    \multicolumn{5}{c}{\textbf{MFNet-M}}
    \\ \hline
    \textbf{Bird} &99.9 &96.5 &98.8 &72.8
    \\ \hline
    \textbf{UAV} &96.5  &94.3 &96.7 &69.9
    \\ \hline
    \textbf{Average} &98.3 &95.4 &97.7 &71.4
    \\ \hline
    \multicolumn{5}{c}{\textbf{MFNet-L}}
    \\ \hline
    \textbf{Bird} &99.6 &97.2 &99.3  &73.9
    \\ \hline
    \textbf{UAV} &97.2 &95.9  &97.3 &69.7
    \\ \hline
    \textbf{Average} &98.4  &96.6 &98.3  &71.8
    \\ \hline  
   \multicolumn{5}{c}{\textbf{(c) Ablation study 3}}
   \\ \hline 
    \textbf{Class} & \textbf{Precision (\%)} & \textbf{Recall (\%)} & \textbf{mAP (\%)}& \textbf{IoU (\%)} 
    \\ \hline
    \multicolumn{5}{c}{\textbf{MFNet-FA-S}}
    \\ \hline
    \textbf{Bird} &96.5 &94.2 &97.5 &65.5
    \\ \hline
    \textbf{UAV} &93.5 &91.9 &93.8  &62.7
    \\ \hline
    \textbf{Average} &95 &93 &95.7 &64.1
    \\ \hline
    \multicolumn{5}{c}{\textbf{MFNet-FA-M}}
    \\ \hline
    \textbf{Bird} &99.8 &93.8 &98.8 &69.3
    \\ \hline
    \textbf{UAV} &93.5 &88.6 &94.4  &66
    \\ \hline
    \textbf{Average} &96.7 &91.2 &96.6  &67.6
    \\ \hline
    \multicolumn{5}{c}{\textbf{MFNet-FA-L}}
    \\ \hline
    \textbf{Bird} &98.9  &97.5  &99.4  &72.7
    \\ \hline
    \textbf{UAV} &95.1 &95.4 &96.8 &71.4
    \\ \hline
    \textbf{Average} &97 &96.5  &98.1 &72.1
    \\ \hline
    \multicolumn{5}{c}{\textbf{(d) YOLOv5s}}
    \\ \hline
    \textbf{Class} & \textbf{Precision (\%)} & \textbf{Recall (\%)} & \textbf{mAP (\%)}& \textbf{IoU (\%)} 
    \\ \hline
    \textbf{Bird} &85.1 &82.5 &85.5  &37.5
    \\ \hline
    \textbf{UAV} &96 &90 &95.6 &61.3
    \\ \hline
    \textbf{Average} &90.5 &86.2 &90.6 & 49.4 
   \\ \hline  
\label{t3}
\vspace{-5mm}
\end{tabular}%
\end{table}

\subsubsection{Ablation study 1}
Ablation study 1 shows the results of a model trained without using the auto batch size and image size selection algorithms. The image size is 416$\times$416, the batch size is 32, and the scale weight decay is 0.0005 for MFNet-S/M/L and YOLOv5s. In Table \ref{t3}, for UAV detection, there is an increasing pattern in precision and recall from MFNet-S to MFNet-M and then a slight decrease in MFNet-L. However, mAP and IoU show a similar increasing pattern from MFNet-S to MFNet-L. Regarding average performance, MFNet-M performs slightly better than MFNet-S, while MFNet-L shows a decrease in precision, recall, and IoU compared to MFNet-M. However, mAP remains relatively stable across all three MFNet models. YOLOv5s performs slightly lower than the MFNet models in precision, recall, and mAP but shows similar IoU values. The performance of MFNet-L is slightly worse than that of MFNet-S/M because of the high false negative rate (birds being detected as drones) and low recall rate. MFNet-S model showcases a reduction of approximately 62.07\% in feature map parameters, an increment of 11.90\% in GFLOPs, and achieves an impressive 18.30\% boost in FPS efficiency with respect to YOLOv5s. MFNet-M displays a 28.97\% feature map parameter reduction accompanied by a substantial GFLOPs increment of approximately 34.56\% w.r.t YOLOv5s with a reduced marginal offset of 4.03\% in FPS efficiency. The MFNet-L model, featuring a 28.97\% decrease in feature map parameters, exhibits a significant surge of 83.19\% in GFLOPs but experiences a notable decline of approximately 41.78\% in FPS efficiency.
For ablation study 1, MFNet-M is the most favorable model for real-time UAV detection due to its superior overall performance, as it achieved 92.3\% precision, 88.4\% recall, 91.5\% mAP, and 51.1\% IoU. That proves that MFNet-M is most sensitive to multi-sized flying birds and other targeted objects in complex background conditions for ablation study 1.

\subsubsection{Ablation study 2}
The outcomes of Ablation Study 2 reveal the performance of a model trained to utilize the automated batch size and image size selection algorithms. Through AutoML, the image size chosen for the dataset is 320$\times$320. DBSA, on the other hand, determined batch sizes and scaled weight decay parameters as follows: 146 batch size and 0.00114 scale weight decay for MFNet-S, 99 batch size and 0.000773 scaled weight decay for MFNet-M, and 73 batch size and 0.000575 scale weight decay for MFNet-L.
In the context of bird detection, a rising trend is observed in precision, recall, mAP, and IoU metrics as we progress from MFNet-S to MFNet-L. However, the metrics values for YOLOv5s are consistently lower compared to the MFNet models. Similarly, in the case of UAV detection, there is a consistent increase in precision, recall, mAP, and IoU metrics from MFNet-S to MFNet-L. Once more, YOLOv5s exhibits lower metrics across the board when compared to the MFNet models. 
The MFNet-S model demonstrates an impressive reduction of 77.77\% in feature map parameters, along with a 2.44\% decrease in GFLOPs. This effective utilization of computational resources results in a substantial 150.48\% increase in FPS efficiency compared to the baseline YOLOv5s. In contrast, MFNet-M experiences a slight divergence with a modest increase of about 4.79\% in feature map parameters, accompanied by a notable GFLOPs surge of 36.39\%. However, there is a minor trade-off as FPS efficiency decreases by roughly 1.90\%. The MFNet-L model introduces a more substantial feature map parameter increase of approximately 48.41\%, followed by a significant rise of about 62.49\% in GFLOPs. Despite these increments, the model maintains a commendable 6.23\% improvement in FPS efficiency.

Considering the overall performance with respect to evaluation metrics, MFNet-L emerges as the best model among all options. It achieves the highest precision, recall, mAP, and IoU values for both bird and UAV detection. YOLOv5s consistently performs lower than the MFNet models in all metrics. MFNet-L is the best model, providing superior bird and UAV detection performance. It demonstrates higher precision, recall, mAP, and IoU values than MFNet-S/M and YOLOv5s as it showed an average precision of 98.4\%, recall of 96.6\%, mAP of 98.3\%, and IoU of 71.8\%. That proves that for Ablation study 2, MFNet-L is most sensitive to multi-sized flying birds and other targeted objects in complex background conditions.
\subsubsection{Ablation study 3}
Ablation study 3 presents the outcomes of the model trained using the proposed feature attention module. With the assistance of AutoML, the image size chosen for the dataset is 320$\times$320. DBSA, in this context, established the following configurations: a batch size of 339 and a scale weight decay of 0.00264 for MFNet-FA-S, a batch size of 160 and a scaled weight decay of 0.00125 for MFNet-FA-M, and a batch size of 112 with a scale weight decay of 0.000875 for MFNet-FA-L. The MFNet-FA-S model showcases an impressive reduction of 83.75\% in feature map parameters and a 29.82\% decrease in GFLOPs, resulting in a remarkable 77.46\% increase in FPS efficiency compared to YOLOv5s. The MFNet-FA-M model displays a reduction of 36.28\% in feature map parameters, a 14.255\% decrease in GFLOPs, and a substantial 71.49\% improvement in FPS efficiency. In contrast, MFNet-FA-L demonstrates a feature map parameter decrease of 25.91\% compared to YOLOv5s. This reduction is offset by a notable GFLOPs increase of approximately 42.49\%, contributing to a commendable 41.04\% enhancement in FPS efficiency.

The MFNet-FA-S model attains an increase of 5.5\% in precision, a 6.5\% rise in the recall, a notable 10.1\% improvement in mAP, and a significant 27\% elevation in IoU compared to YOLOv5s. On the other hand, the MFNet-FA-M model showcases a precision boost of 6.2\%, a higher recall by 5.2\%, an enhanced mAP by 6\%, and a substantial 17.3\% surge in IoU. The MFNet-FA-L model demonstrates significant progress over YOLOv5s, featuring a remarkable 12.9\% increase in precision, a substantial 14.3\% enhancement in the recall, an mAP boost of approximately 7.5\%, and an exceptional 33.6\% elevation in IoU. In combination, these evaluation metrics highlight the clear superiority of the MFNet-FA models over YOLOv5s. Consequently, the MFNet-FA model presents improved object detection accuracy and localization capabilities, rendering it the preferred option for tackling the drone versus bird detection challenge.

\subsection{Comparison of Ablation Studies: Feature map sizes, computational resources, and efficiency}
 Table \ref{t2} provides a comprehensive breakdown of the total extracted feature map parameters/sizes (TEFS), computational resources (CR) expressed in terms of GFLOPs, and efficiency analysis in relation to FPS rates for all ablation studies compared to YOLOv5s.

In Ablation Study 1, the MFNet-S model emerges as the optimal choice due to its substantial 62.07\% reduction in feature map parameters compared to YOLOv5s. It maintains competitive FPS efficiency, with a noteworthy 18.30\% improvement over YOLOv5s, while only slightly increasing GFLOPs by 11.90\% compared to YOLOv5s. Moving to Ablation study 2, MFNet-S maintains its superiority by featuring significantly reduced feature map parameters (77.77\% decrease compared to YOLOv5s), along with an enhanced FPS efficiency (50.48\% improvement over YOLOv5s), and a 2.44\% decrease in GFLOPs relative to YOLOv5s. In Ablation study 3, the MFNet-FA-L model showcases a well-balanced performance profile. It demonstrates a manageable reduction in feature map parameters (25.91\% decrease compared to YOLOv5s), while maintaining competitive FPS efficiency (41.04\% improvement over YOLOv5s) and GFLOPs consumption (42.249\% increase over YOLOv5s).
These statistics show that MFNet-FA-L is the preferred choice for real-time drone versus bird detection, due to its well-rounded performance in terms of feature map parameters, FPS efficiency, and GFLOPs utilization.
  
\begin{table}[!b]
\caption{Ablation studies comparative analysis: extracted feature map parameters, computational resources, and efficiency.}
\centering
    \begin{tabular}{p{1.75cm}p{2.25cm}p{2cm}p{1.25cm}}
    \hline
    \textbf{Model} &\textbf{Extracted feature map parameters (Million)} & \textbf{Computational resources (GFLOPs)} & \textbf{Efficiency (FPS)} 
    \\ \hline
       \multicolumn{4}{c}{\textbf{(1) Ablation study 1}}
        % Backbone
        \\ \hline
        \textbf{MFNet-S} &2.7	&18.9 &114.95
        \\ \hline
        \textbf{MFNet-M}&5.2	&75.3 &101.01	
        \\ \hline
        \textbf{MFNet-L}&10.2	&157.6 &56.49
        \\ \hline
        \multicolumn{4}{c}{\textbf{(b) Ablation study 2}}
        \\ \hline
        \textbf{MFNet-S} &1.61 &17.3 &243.90 
        \\ \hline
        \textbf{MFNet-M} &6.9 &78.2  &95.23 
        \\ \hline
        \textbf{MFNet-L} &10.8 &122.2 &103.09
        \\ \hline
        \multicolumn{4}{c}{\textbf{(b) Ablation study 3}}
        \\ \hline
        \textbf{MFNet-FA-S} &1.18 &11.9 &172.4
        \\ \hline
        \textbf{MFNet-FA-M} &4.60 &41.1 &166.6
        \\ \hline
        \textbf{MFNet-FA-L} &10.26 &88.3 &136.98
        \\ \hline
        \multicolumn{4}{c}{\textbf{YOLOv5s}}
         \\ \hline
        \textbf{YOLOv5s}  &7.25 &16.9 &97.08
       \\\hline
    \label{t2}
       \end{tabular}
\end{table}

\subsection{Image attributes affecting detection}
We conduct an assessment of crucial image attributes, including environmental backgrounds, target scales, and other intricate conditions, to gauge their impact on detection performance. The effectiveness of MFNet is influenced by a range of factors, such as training inadequacies and varying parameters. To ensure a comprehensive evaluation, we utilize the detection accuracy of these algorithms as the benchmark for fair assessment.
\begin{figure}[!ht]
     \centering
     \begin{subfigure}[b]{0.07\textwidth}
         \centering
         \includegraphics[width=\textwidth]{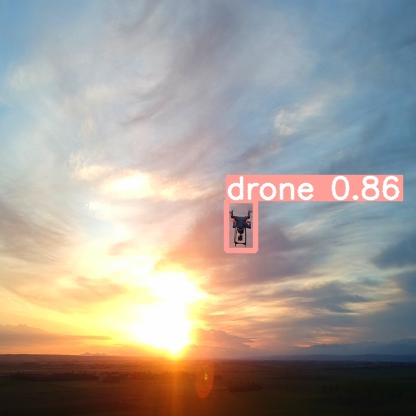}
         \includegraphics[width=\textwidth]{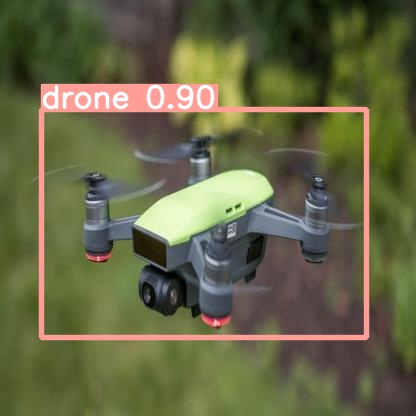}
         \includegraphics[width=\textwidth]{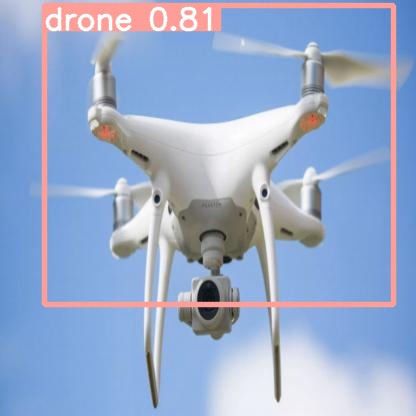}
         \includegraphics[width=\textwidth]{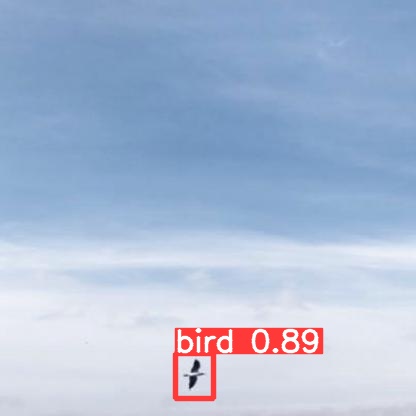}
         \includegraphics[width=\textwidth]{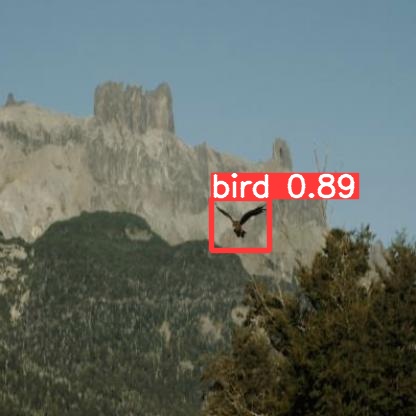}
         \includegraphics[width=\textwidth]{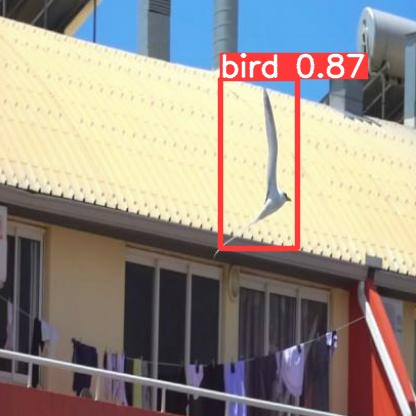}
         \caption{YOLOv5s}
         \label{figa}
     \end{subfigure}
     \hfill
     \begin{subfigure}[b]{0.07\textwidth}
         \centering
         \includegraphics[width=\textwidth]{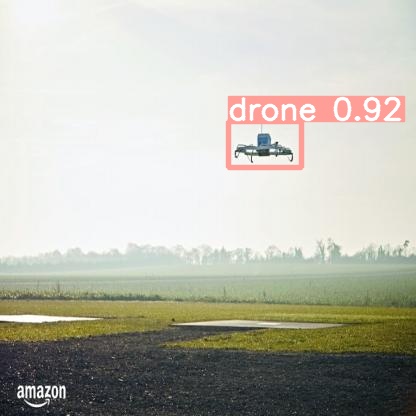}
         \includegraphics[width=\textwidth]{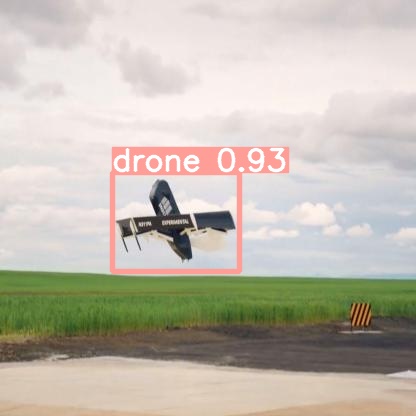}
         \includegraphics[width=\textwidth]{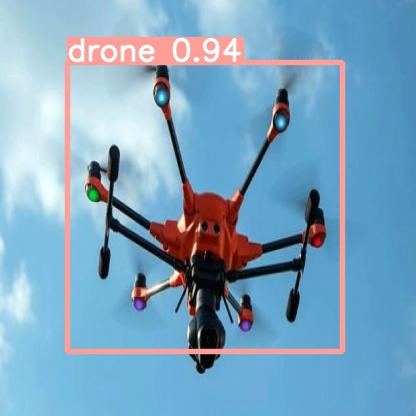}
         \includegraphics[width=\textwidth]{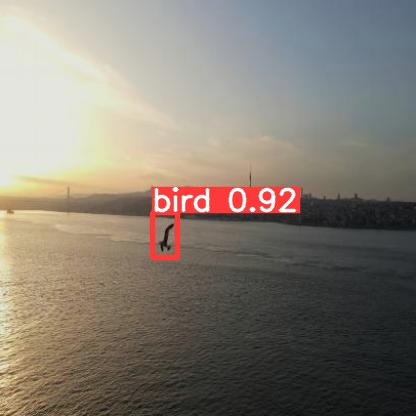}
         \includegraphics[width=\textwidth]{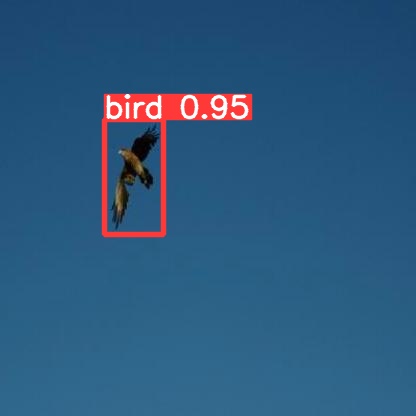}
         \includegraphics[width=\textwidth]{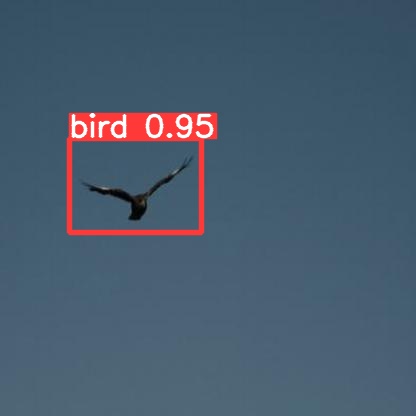}
         \caption{MFNet-S}
         \label{figb}
     \end{subfigure}
     \hfill
     \begin{subfigure}[b]{0.07\textwidth}
         \centering
        \includegraphics[width=\textwidth]{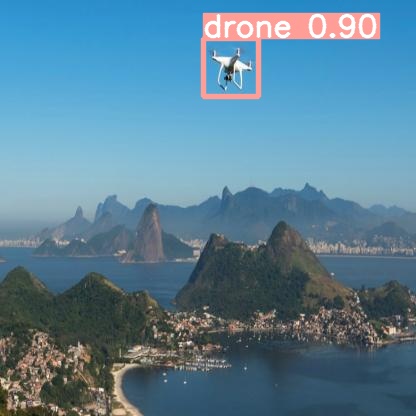}
         \includegraphics[width=\textwidth]{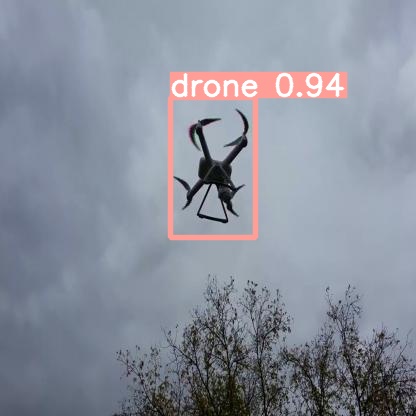}
         \includegraphics[width=\textwidth]{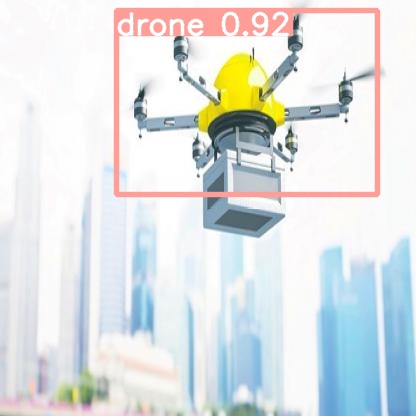}
         \includegraphics[width=\textwidth]{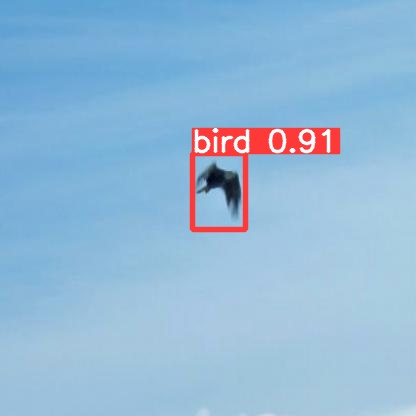}
         \includegraphics[width=\textwidth]{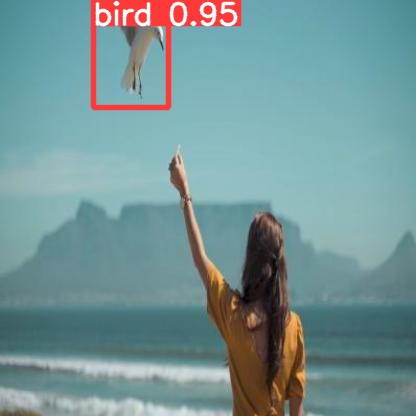}
         \includegraphics[width=\textwidth]{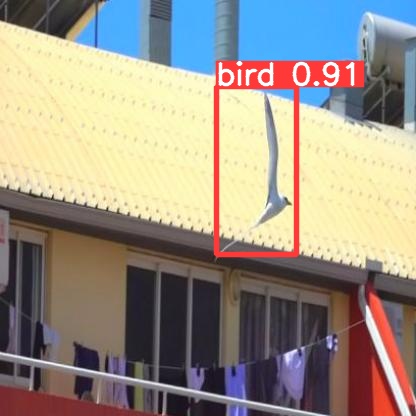}
         \caption{MFNetM}
         \label{figc}
     \end{subfigure}
     \hfill
     \begin{subfigure}[b]{0.07\textwidth}
         \centering
         \includegraphics[width=\textwidth]{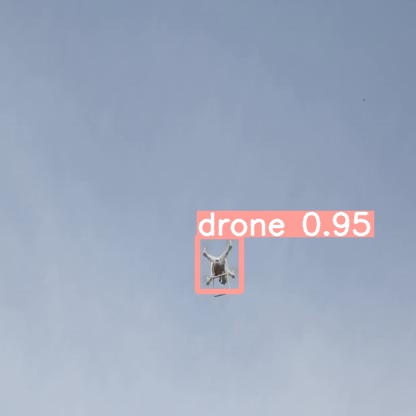}
         \includegraphics[width=\textwidth]{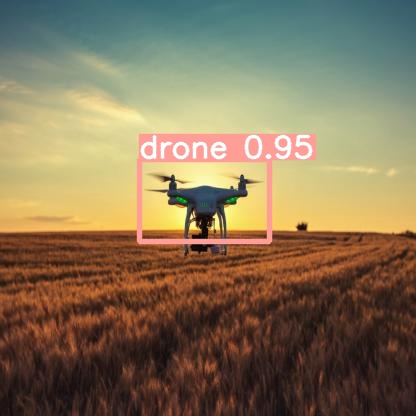}
         \includegraphics[width=\textwidth]{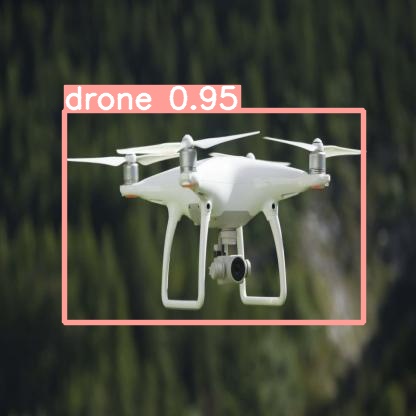}
         \includegraphics[width=\textwidth]{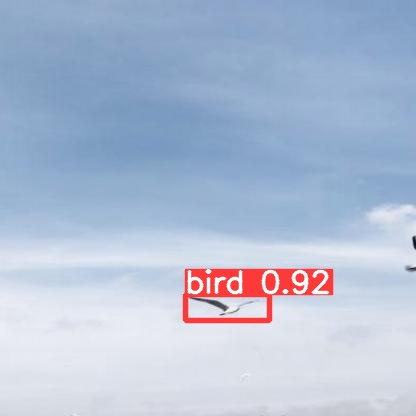}
         \includegraphics[width=\textwidth]{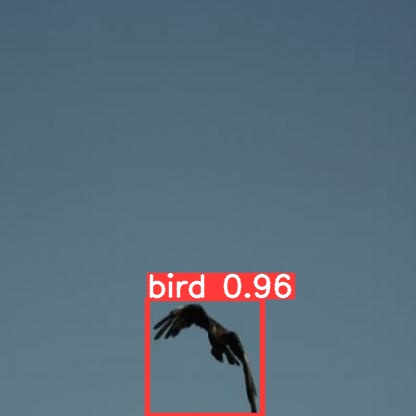}
         \includegraphics[width=\textwidth]{mfnetm_large_drone.jpg}
         \caption{MFNet-L}
         \label{figd}
     \end{subfigure}
     \hfill
     \caption{Ablation study 2: Detection results with respect to target scale (Top to bottom) for small-sized, medium-sized, and large-sized drones and birds.}
        \label{f5}
\end{figure}
\subsubsection{Target scales}
The image size of the target, whether small, medium, or large, plays a significant role in shaping the model's detection performance. We depict the detection outcomes for UAVs and birds across all selected models in Fig. \ref{f5} and Fig. \ref{f6}. The results underscore that MFNet-L achieves the highest detection accuracy (95 \%) across all sizes of UAV targets. For small-sized bird targets, both MFNet-S and MFNet-L achieve the highest detection accuracy of 92 \%. In the case of medium-sized bird targets, MFNet-L demonstrates the highest detection accuracy at 96\%, while both MFNet-S and MFNet-L achieve equally strong scores (95\%) for large-sized bird targets. Therefore, the MFNet-L model with the focus module, as observed in Ablation study 2, emerges as an exceptional selection for detecting individual bird and drone targets within frames of varying sizes, even under challenging circumstances. This reinforces the notion that integrating the focus module contributes to the process of narrowing down and concentrating on specific targets.

\subsubsection{Impact of varying environment backgrounds on detection}
\begin{figure}[!t]
     \centering
     \begin{subfigure}[b]{0.07\textwidth}
         \centering
         \includegraphics[width=\textwidth]{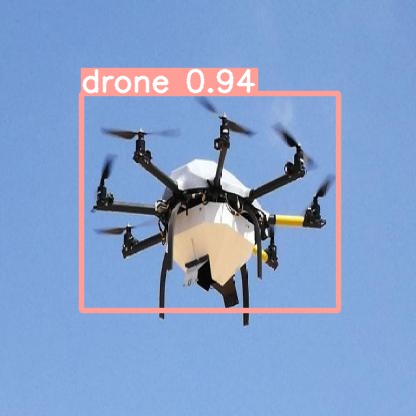}
         \includegraphics[width=\textwidth]{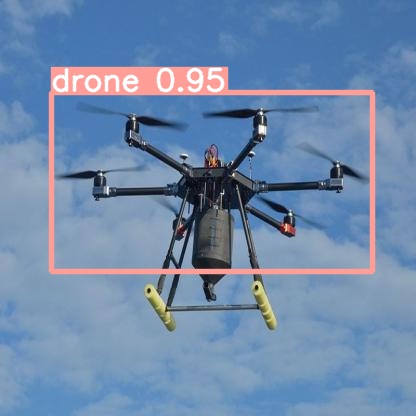}
         \includegraphics[width=\textwidth]{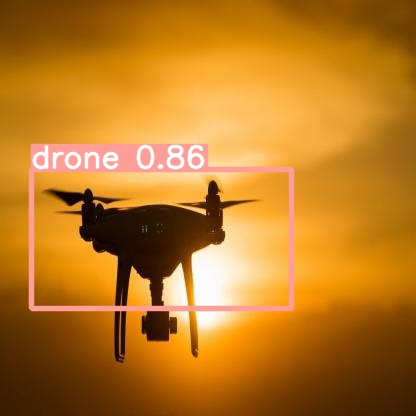}
         \includegraphics[width=\textwidth]{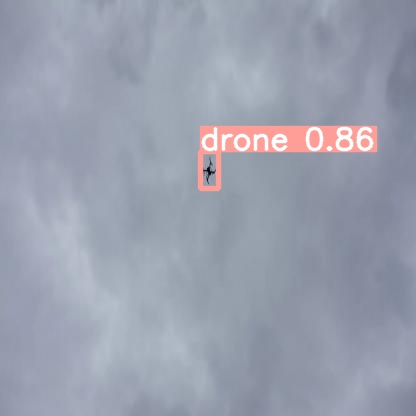}
         \includegraphics[width=\textwidth]{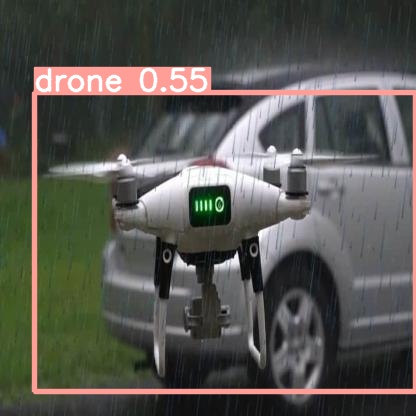}
         \includegraphics[width=\textwidth]{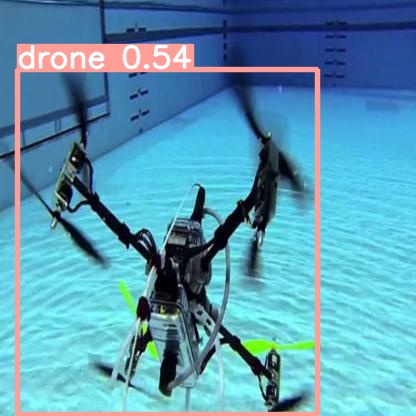}
         \includegraphics[width=\textwidth]{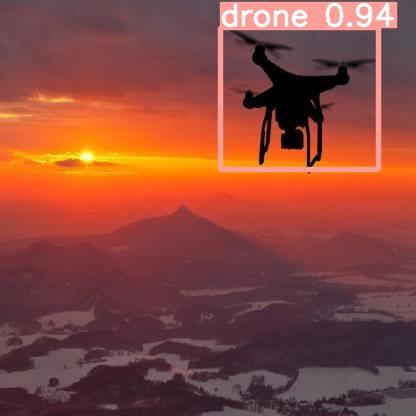}
         \includegraphics[width=\textwidth]{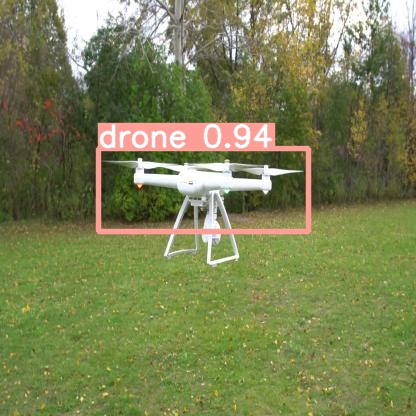}
         \caption{YOLOv5s}
         \label{figa}
     \end{subfigure}
     \hfill
     \begin{subfigure}[b]{0.07\textwidth}
         \centering
         \includegraphics[width=\textwidth]{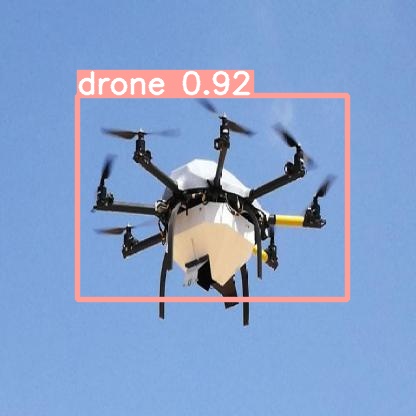}
         \includegraphics[width=\textwidth]{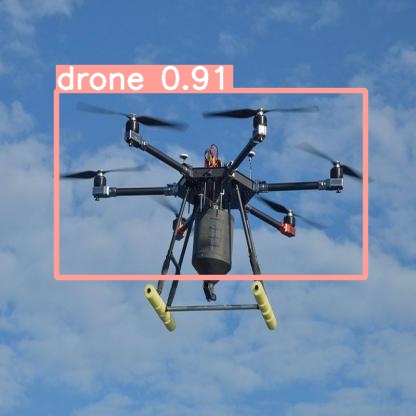}
         \includegraphics[width=\textwidth]{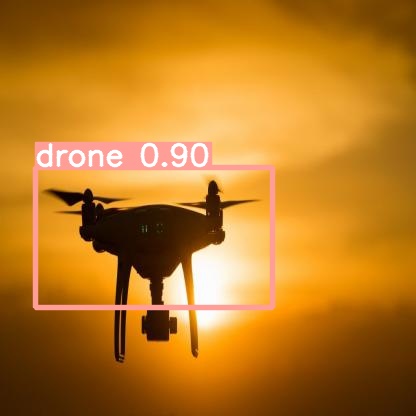}
         \includegraphics[width=\textwidth]{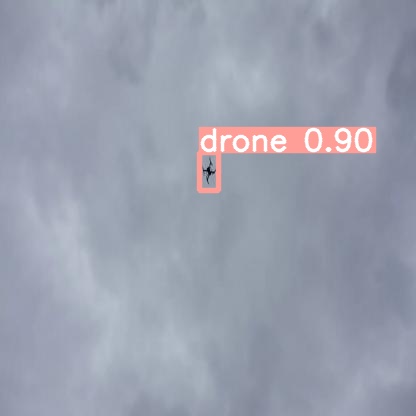}
         \includegraphics[width=\textwidth]{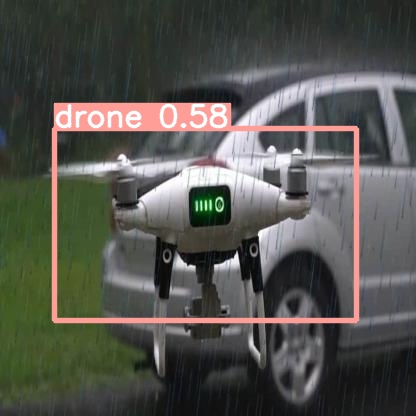}
         \includegraphics[width=\textwidth]{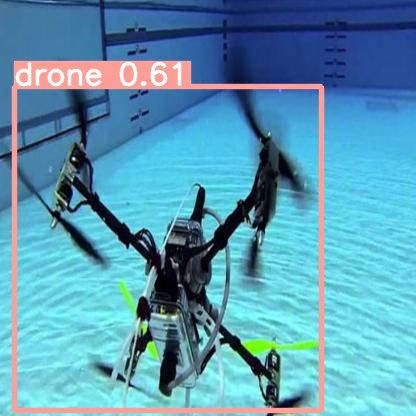}
         \includegraphics[width=\textwidth]{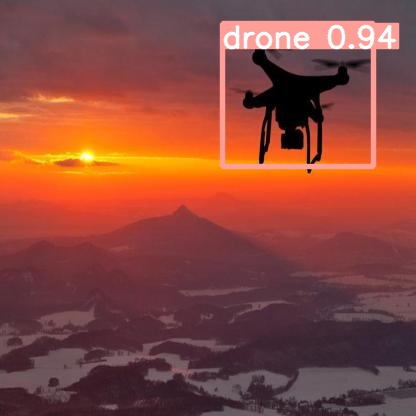}
         \includegraphics[width=\textwidth]{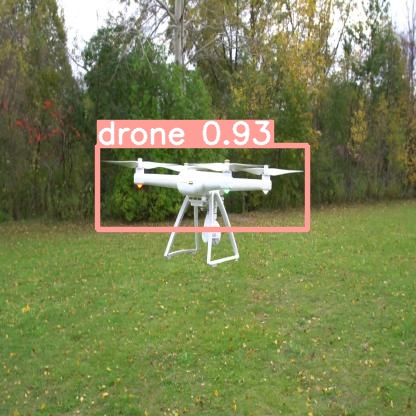}
         \caption{MFNet-S}
         \label{figb}
     \end{subfigure}
     \hfill
     \begin{subfigure}[b]{0.07\textwidth}
         \centering
         \includegraphics[width=\textwidth]{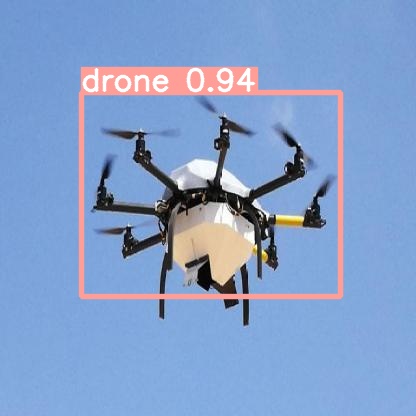}
         \includegraphics[width=\textwidth]{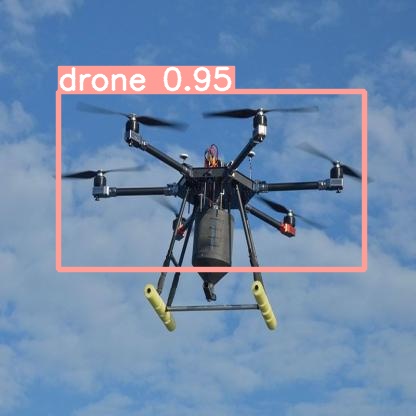}
         \includegraphics[width=\textwidth]{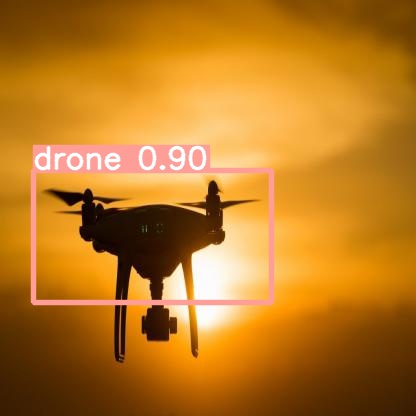}
         \includegraphics[width=\textwidth]{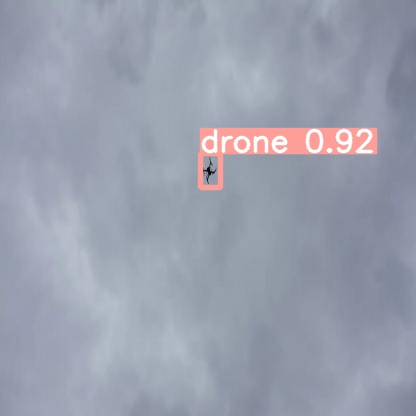}
         \includegraphics[width=\textwidth]{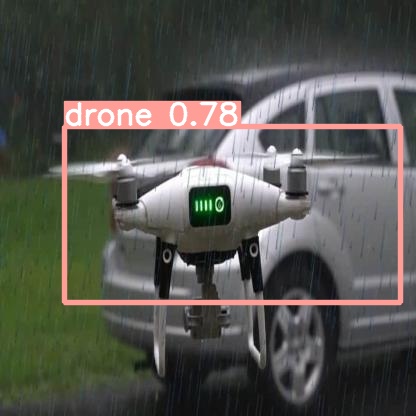}
         \includegraphics[width=\textwidth]{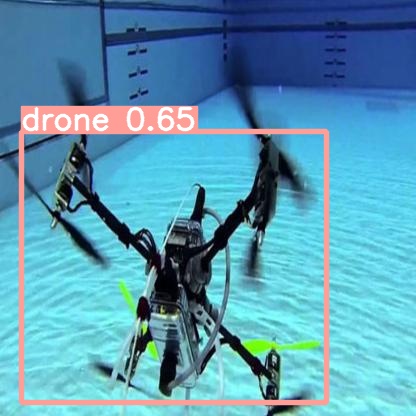}
         \includegraphics[width=\textwidth]{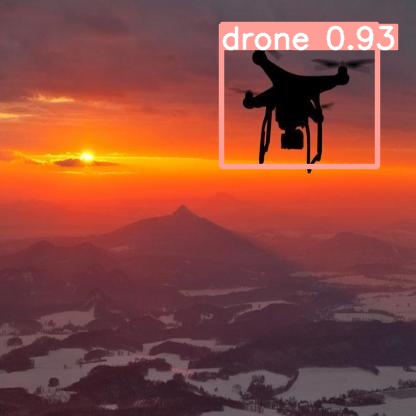}
         \includegraphics[width=\textwidth]{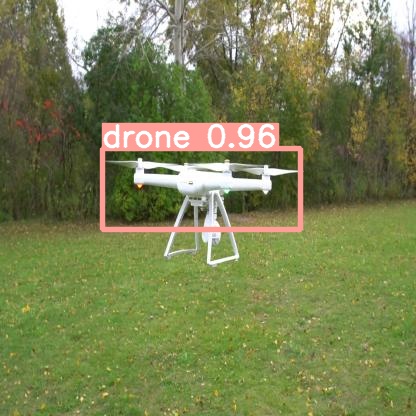}
         \caption{MFNetM}
         \label{figc}
     \end{subfigure}
     \hfill
     \begin{subfigure}[b]{0.07\textwidth}
         \centering
         \includegraphics[width=\textwidth]{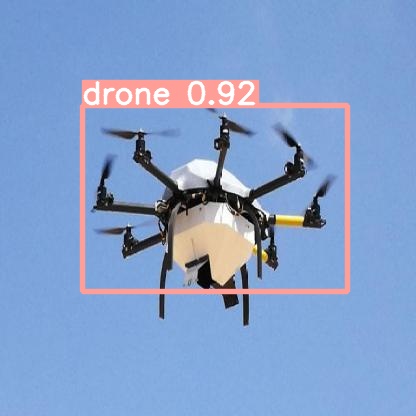}
         \includegraphics[width=\textwidth]{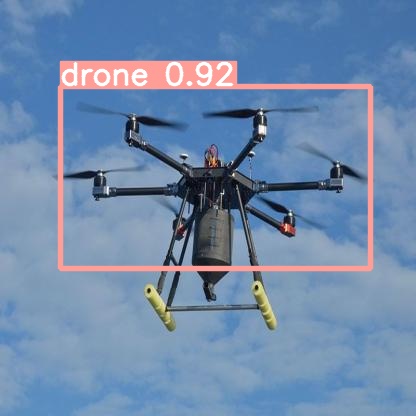}
         \includegraphics[width=\textwidth]{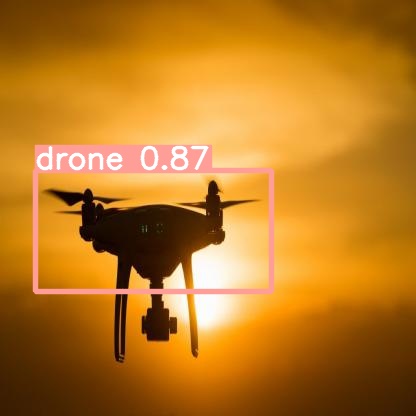}
         \includegraphics[width=\textwidth]{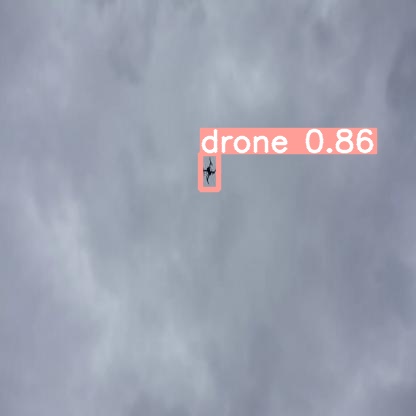}
         \includegraphics[width=\textwidth]{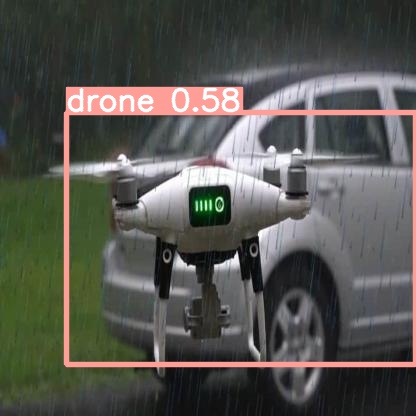}
         \includegraphics[width=\textwidth]{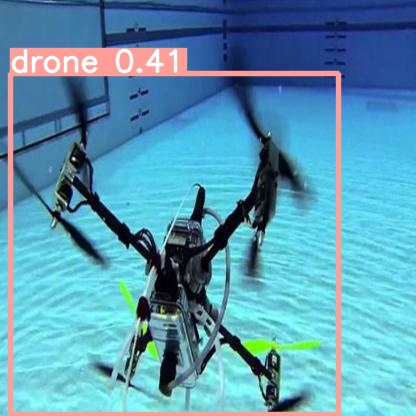}
         \includegraphics[width=\textwidth]{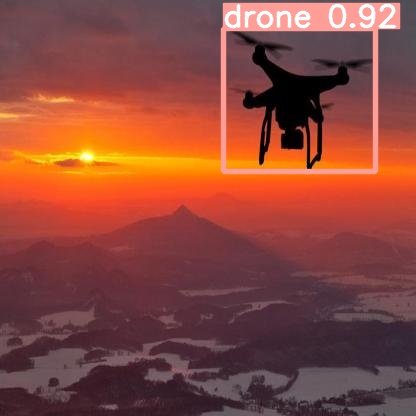}
         \includegraphics[width=\textwidth]{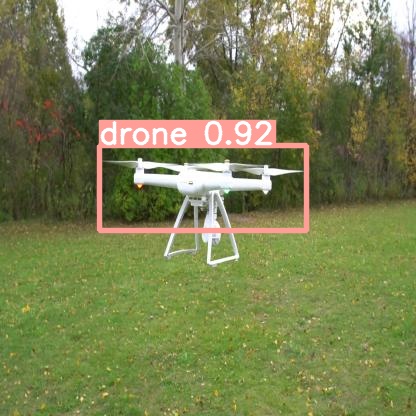}
         \caption{MFNet-L}
         \label{figd}
     \end{subfigure}
     \hfill
     \caption{Ablation study 2: UAV detection performance for proposed MFNet and baseline under challenging environmental conditions (Top to bottom) like clear sky, cloudy conditions, sunlight, foggy weather, Rainy situation, waterfall, mountains, and forest.}
        \label{f6}
\end{figure}

\begin{figure}[!ht]
     \centering
     \begin{subfigure}[b]{0.07\textwidth}
         \centering
         \includegraphics[width=\textwidth]{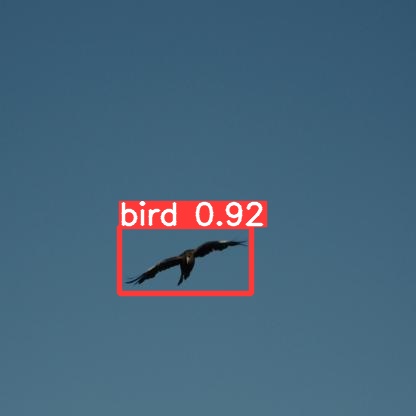}
         \includegraphics[width=\textwidth]{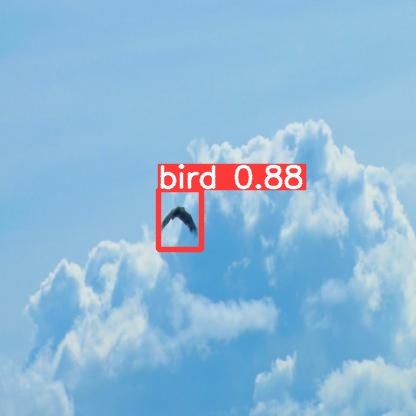}
         \includegraphics[width=\textwidth]{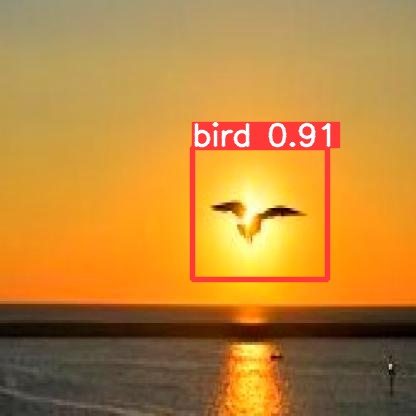}
         \includegraphics[width=\textwidth]{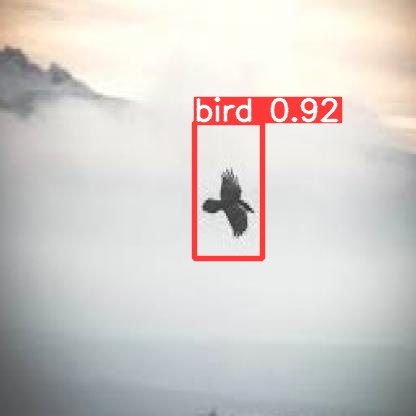}
         \includegraphics[width=\textwidth]{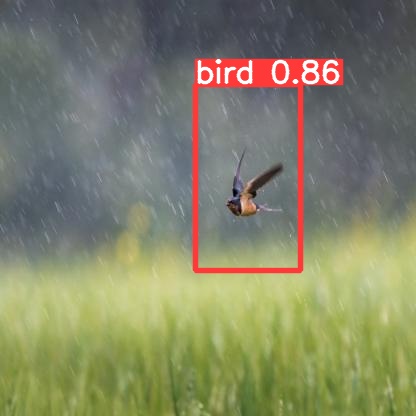}
         \includegraphics[width=\textwidth]{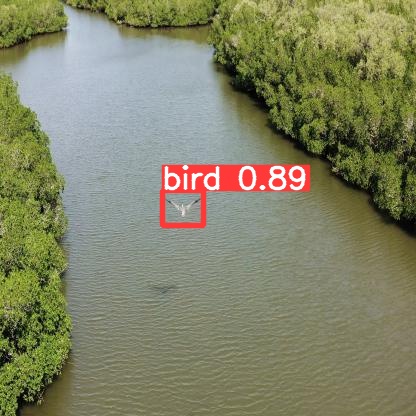}
         \includegraphics[width=\textwidth]{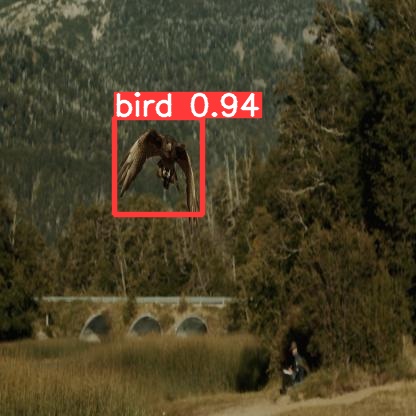}
         \caption{YOLOv5s}
         \label{figa}
     \end{subfigure}
     \hfill
     \begin{subfigure}[b]{0.07\textwidth}
         \centering
         \includegraphics[width=\textwidth]{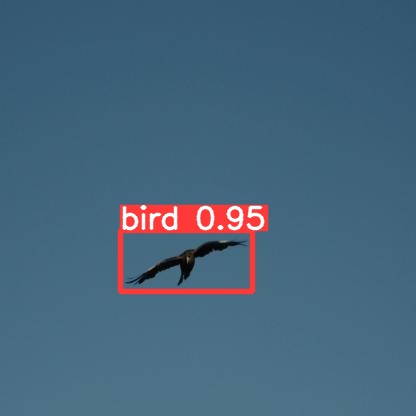}
         \includegraphics[width=\textwidth]{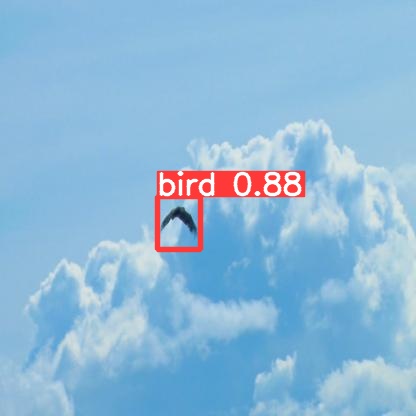}
         \includegraphics[width=\textwidth]{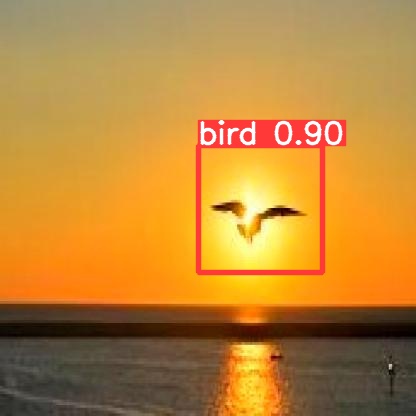}
         \includegraphics[width=\textwidth]{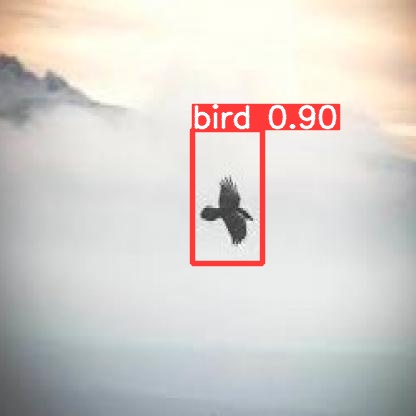}
         \includegraphics[width=\textwidth]{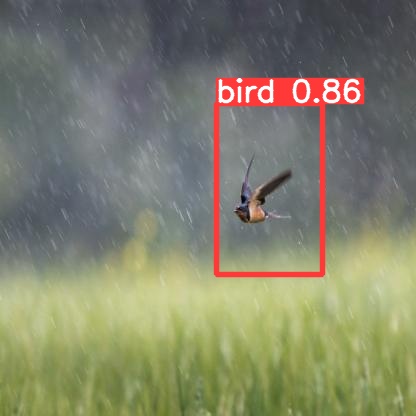}
         \includegraphics[width=\textwidth]{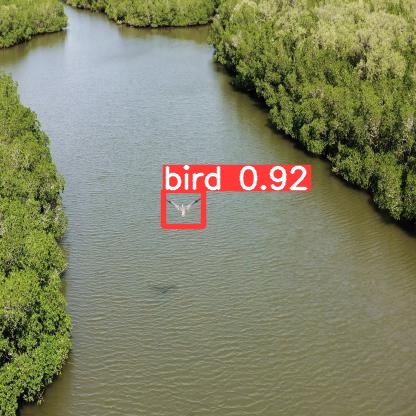}
         \includegraphics[width=\textwidth]{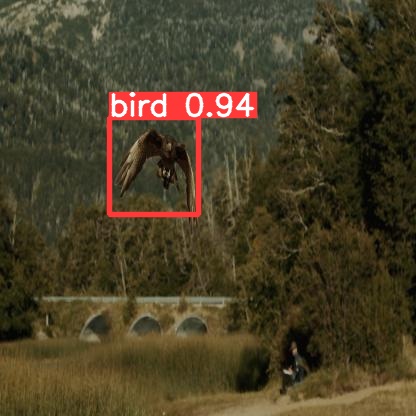}
         \caption{MFNet-S}
         \label{figb}
     \end{subfigure}
     \hfill
     \begin{subfigure}[b]{0.07\textwidth}
         \centering
         \includegraphics[width=\textwidth]{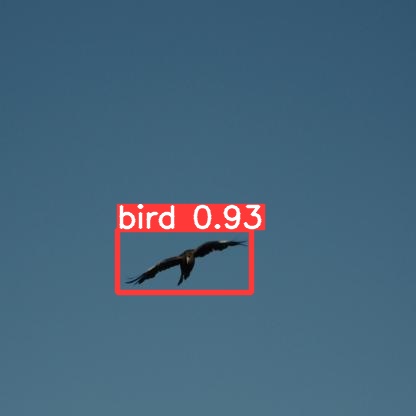}
         \includegraphics[width=\textwidth]{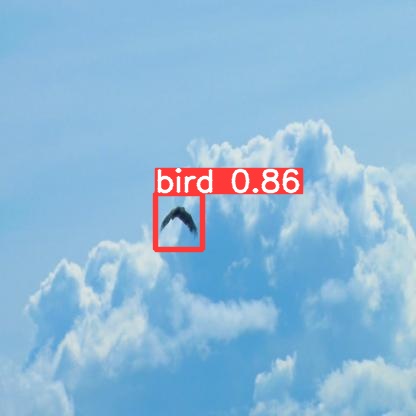}
         \includegraphics[width=\textwidth]{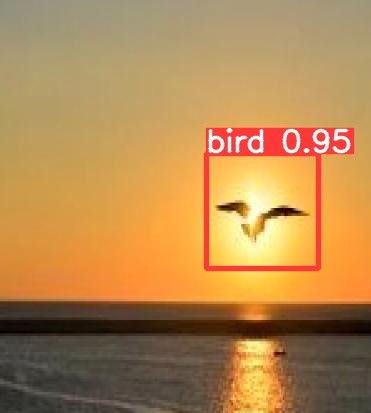}
         \includegraphics[width=\textwidth]{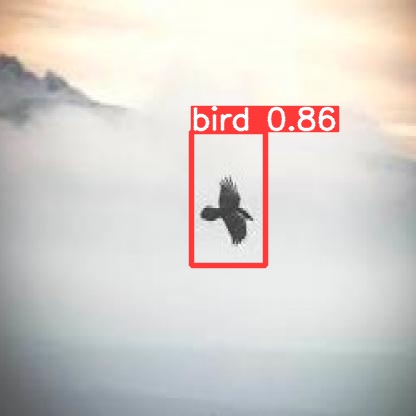}
         \includegraphics[width=\textwidth]{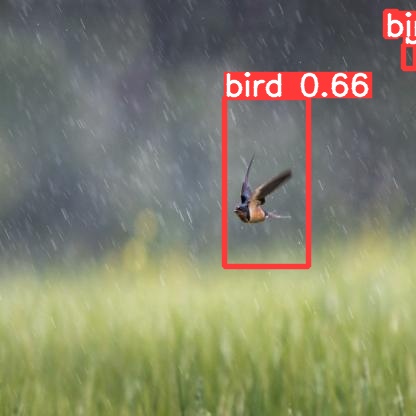}
         \includegraphics[width=\textwidth]{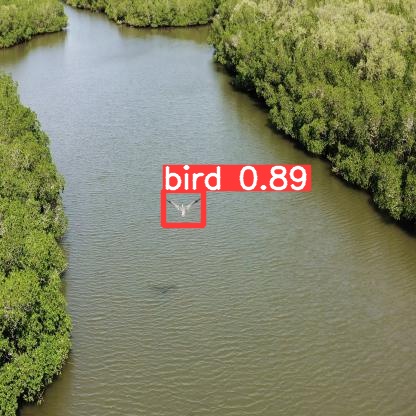}
         \includegraphics[width=\textwidth]{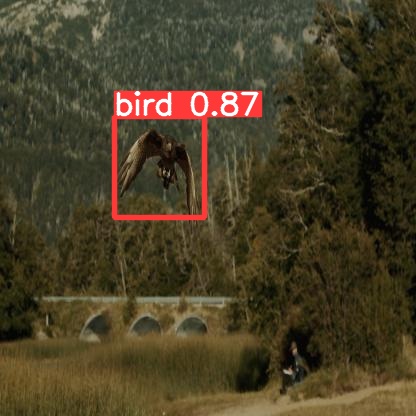}
         \caption{MFNet-M}
         \label{figc}
     \end{subfigure}
     \hfill
     \begin{subfigure}[b]{0.07\textwidth}
         \centering
        \includegraphics[width=\textwidth]{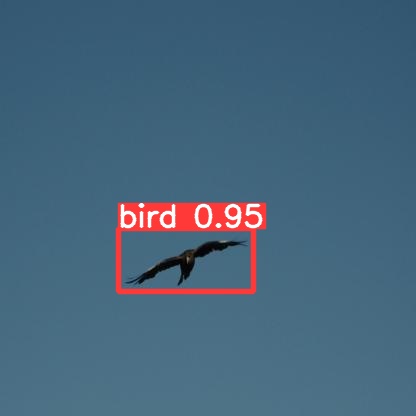}
         \includegraphics[width=\textwidth]{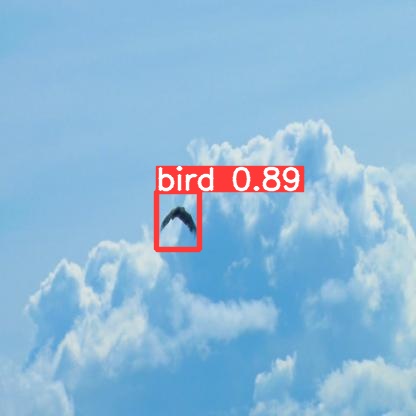}
         \includegraphics[width=\textwidth]{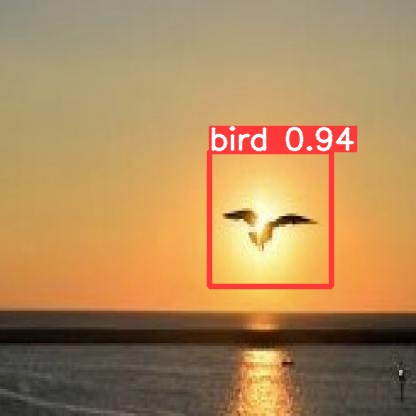}
         \includegraphics[width=\textwidth]{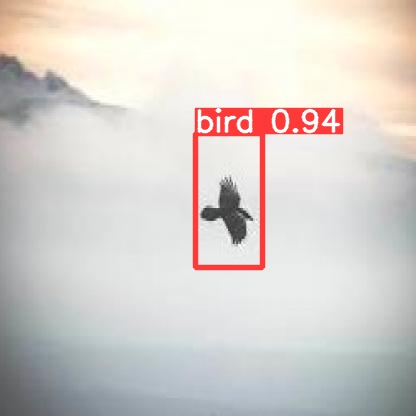}
         \includegraphics[width=\textwidth]{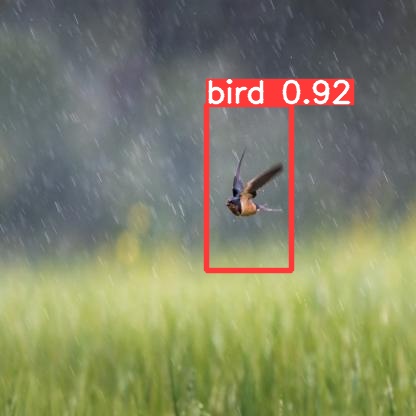}
         \includegraphics[width=\textwidth]{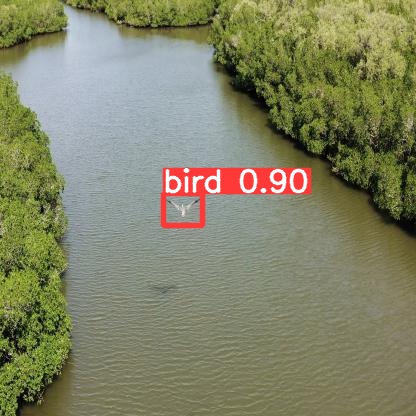}
         \includegraphics[width=\textwidth]{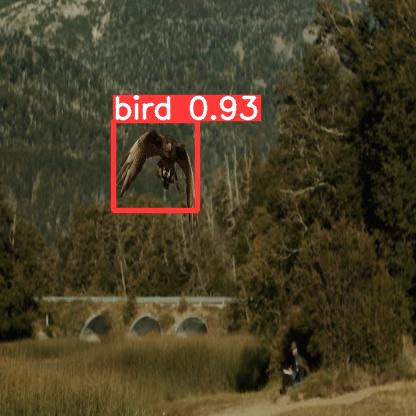}
         \caption{MFNet-L}
         \label{figd}
     \end{subfigure}
     \hfill
     \caption{Ablation study 2: Bird detection performance for proposed MFNet-FA and baseline under challenging environmental conditions (Top to bottom) like clear sky, cloudy conditions, sunlight, foggy weather, Rainy situation, waterfall, mountains, and forest.}
        \label{f7}
\end{figure}

In Figure \ref{f6} and Figure \ref{f7}, we present a comparison of the detection performance of the proposed MFNet/MFNet-FA under different challenging background conditions for UAVs and birds, respectively. In this analysis, due to superior results, we focus solely on discussing the performance of MFNet in comparison to YOLOv5s.
Both YOLOv5s and MFNet-L achieve the highest detection accuracy of 94\% for UAV images captured under clear skies and cloudy weather conditions. Similarly, for UAV images taken in sunny conditions, MFNet-S and MFNet-L attain a peak detection accuracy of 90\%. MFNet-L demonstrates a detection accuracy of 92\% for UAVs against foggy backgrounds, 78\% for UAVs in rainy situations, 65\% for UAVs flying over water backgrounds, and an impressive 96\% for UAVs within green forest settings. Furthermore, for UAVs flying in hilly areas, both YOLOv5s and MFNet-M achieve a detection accuracy of 94\%.

In the case of bird detection, MFNet-S and MFNet-L perform exceptionally well, achieving a remarkable 95\% accuracy under clear sky conditions. Meanwhile, YOLOv5s and MFNet-L manage around 86\% detection accuracy for rainy and hilly areas, and 94\% accuracy in forested environments. Moreover, MFNet-L demonstrates its proficiency by achieving an accuracy of 89\% under cloudy sky conditions, 94\% in sunny weather, and 94\% in foggy conditions. We conclude that the rainy condition and the cloudy plus rainy conditions are the most challenging for UAV and bird detection, respectively.

\subsubsection{Multiple targets detection in a single scene with challenging conditions}

\begin{figure}[!h]
     \centering
     \begin{subfigure}[b]{0.07\textwidth}
         \centering
         \includegraphics[width=\textwidth]{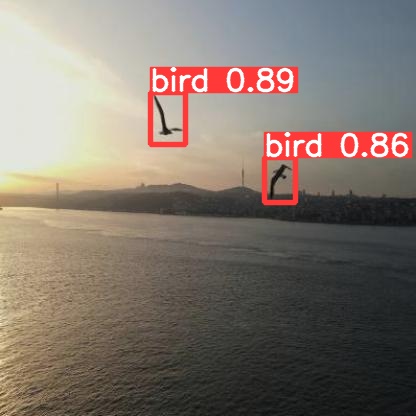}
         \includegraphics[width=\textwidth]{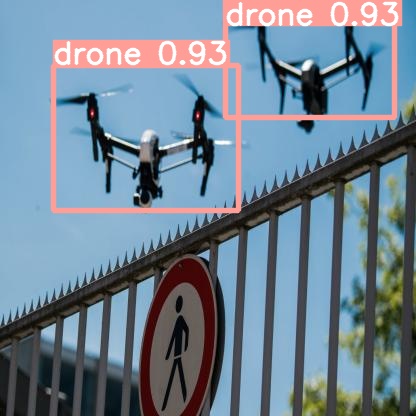}
         \includegraphics[width=\textwidth]{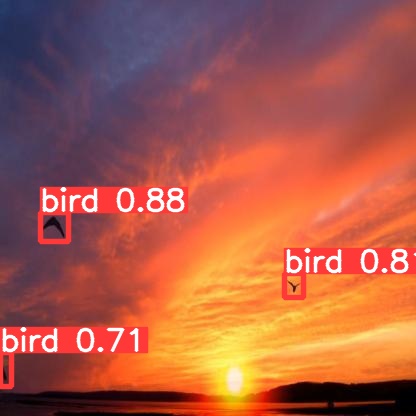}
         \includegraphics[width=\textwidth]{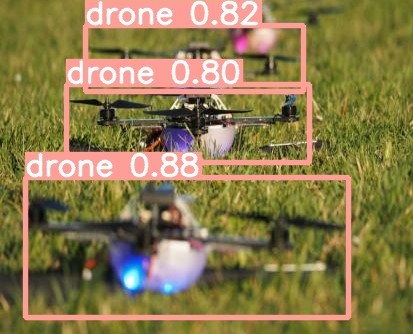}
         \includegraphics[width=\textwidth]{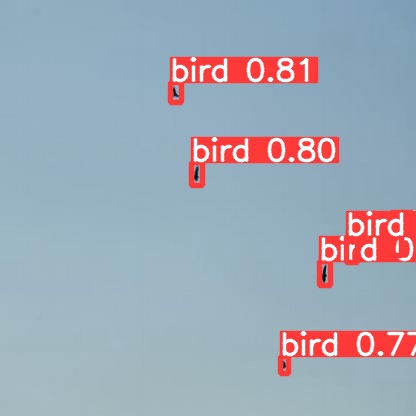}
         \includegraphics[width=\textwidth]{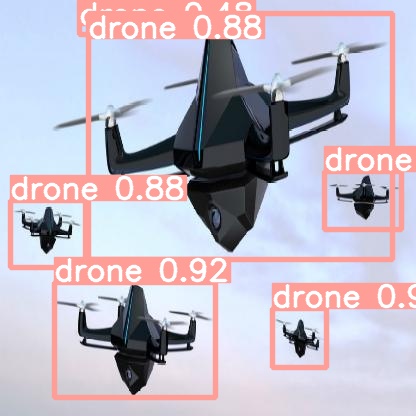}
         \includegraphics[width=\textwidth]{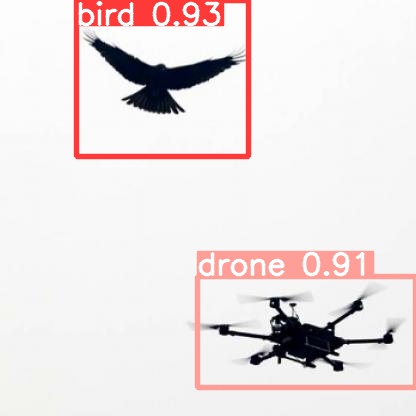}
         \caption{YOLOv5s}
         \label{figa}
     \end{subfigure}
     \hfill
     \begin{subfigure}[b]{0.07\textwidth}
         \centering
         \includegraphics[width=\textwidth]{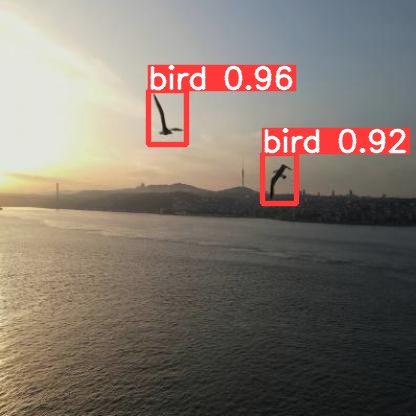}
         \includegraphics[width=\textwidth]{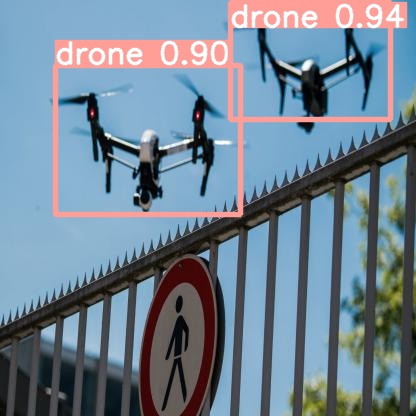}
         \includegraphics[width=\textwidth]{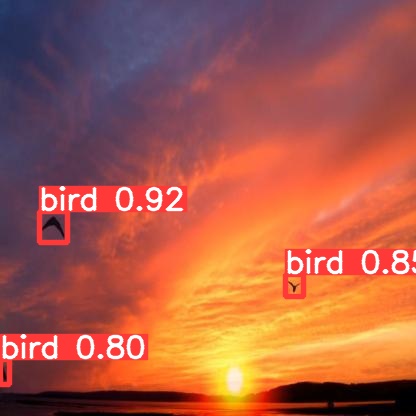}
         \includegraphics[width=\textwidth]{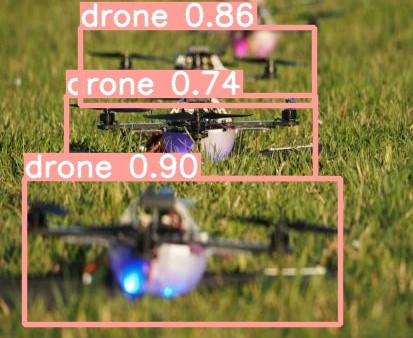}
         \includegraphics[width=\textwidth]{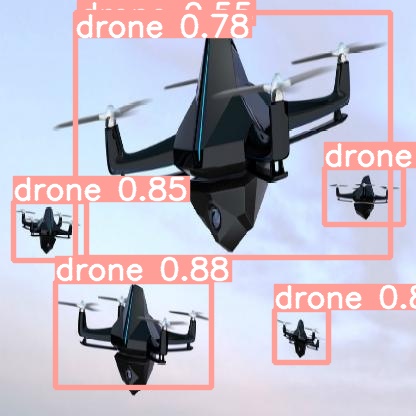}
         \includegraphics[width=\textwidth]{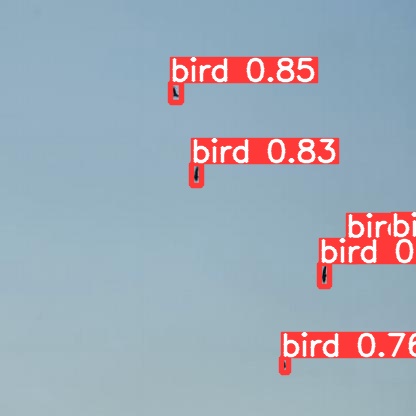}
         \includegraphics[width=\textwidth]{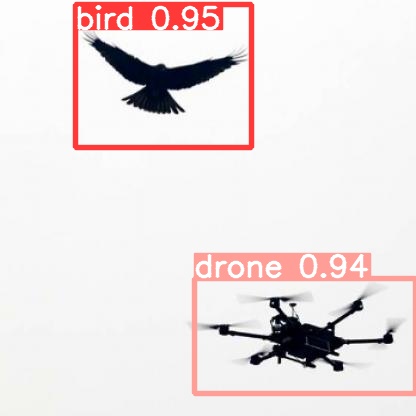}
         \caption{MFNet-FA-S}
         \label{figb}
     \end{subfigure}
     \hfill
     \begin{subfigure}[b]{0.07\textwidth}
        \includegraphics[width=\textwidth]{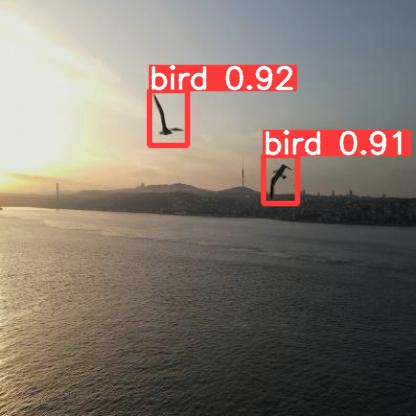}
         \includegraphics[width=\textwidth]{mfnetl2birds.jpg}
         \includegraphics[width=\textwidth]{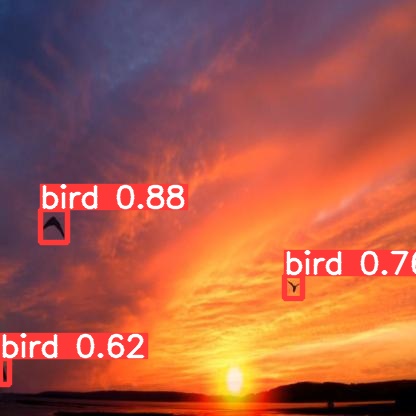}
         \includegraphics[width=\textwidth]{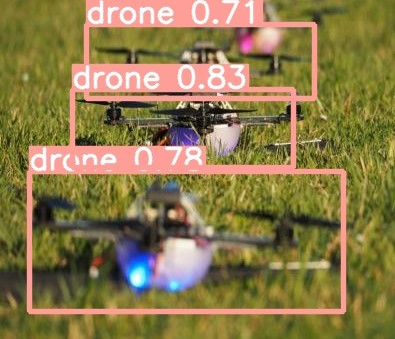}
         \includegraphics[width=\textwidth]{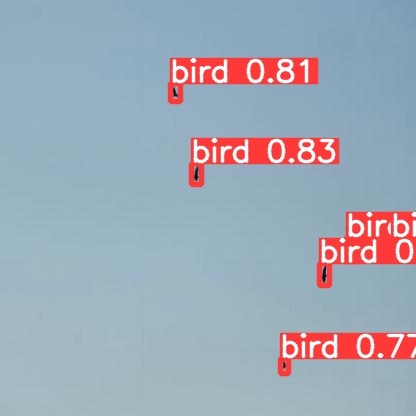}
         \includegraphics[width=\textwidth]{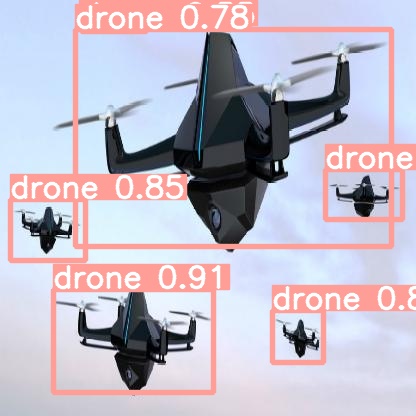}
         \includegraphics[width=\textwidth]{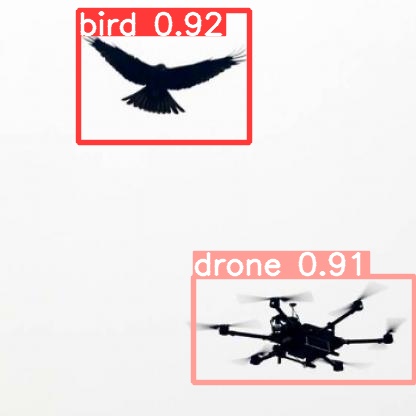}
         \caption{MFNet-FA-M}
         \label{figc}
     \end{subfigure}
     \hfill
     \begin{subfigure}[b]{0.07\textwidth}
         \centering
         \includegraphics[width=\textwidth]{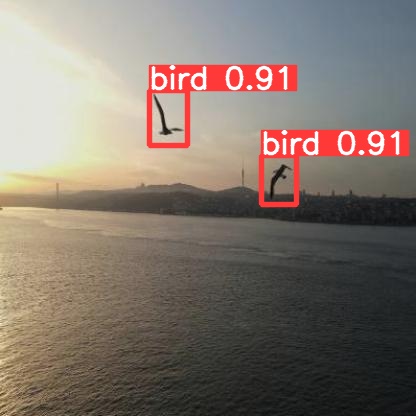}
         \includegraphics[width=\textwidth]{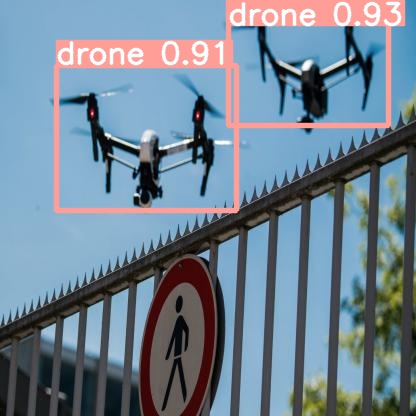}
         \includegraphics[width=\textwidth]{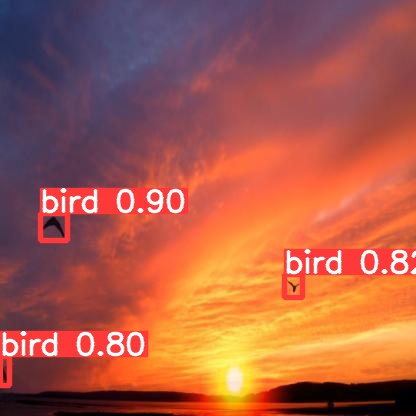}
         \includegraphics[width=\textwidth]{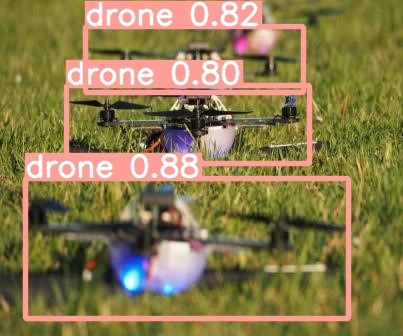}
         \includegraphics[width=\textwidth]{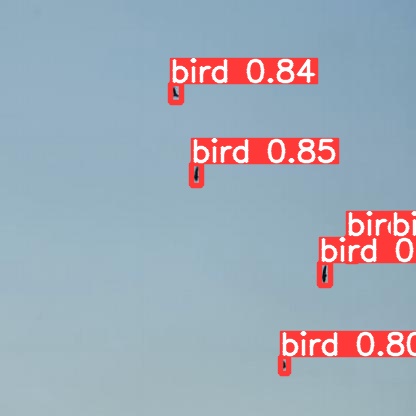}
         \includegraphics[width=\textwidth]{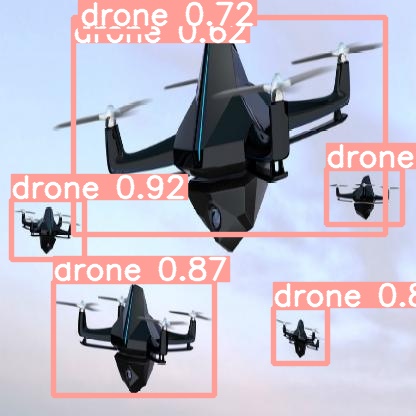}
         \includegraphics[width=\textwidth]{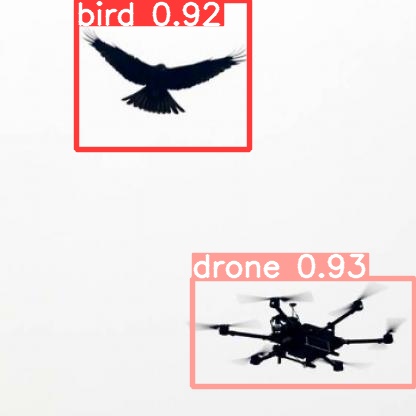}
         \caption{MFNet-FA-L}
         \label{figd}
     \end{subfigure}
     \hfill
     \caption{Ablation study 3 (MFNet-FA): Multiple-targets (birds or drones) detection performance in a single scene with challenging environmental conditions.}
        \label{f8}
 \end{figure}

The growing attention towards multiple UAV swarms has captivated the interests of domestic, commercial, and military users, primarily due to their potential to enhance performance. Nonetheless, the deployment of these swarms mandates robust collision avoidance and detection technologies, especially in congested airspace. This emphasizes the pivotal role of high-precision multi-target detection under challenging conditions. We assess the performance of the proposed MFNet/MFNet-FA in the context of multi-target scenarios set within challenging environments. Among all ablation studies, MFNet-FA emerges with the best results, which is why we exclusively present its comparison to YOLOv5s in Fig. \ref{f8}. 

In scenarios where an image contains 2 birds and 2 UAVs, MFNet-FA-S exhibited the highest level of detection accuracy. This pattern holds true for instances with 3 birds and UAVs within the image as well. Conversely, when dealing with 5 birds and UAVs present in an image, MFNet-FA-L demonstrated the most accurate detection. Moreover, when tasked with identifying individual birds and UAVs in a single image, MFNet-FA-S showcased the most superior performance. These results indicate the effectiveness of the feature attention module in recognizing multiple objects of varying sizes within images. By generating focused feature maps, this module significantly contributes to enhancing detection accuracy.
\vspace{-1mm}
\subsection{Computational complexity and inference time for UAV detection}
Achieving a high inference rate is pivotal for real-time system deployment and enhancing the speed of UAV detection. It's important to note that commercial UAVs can attain speeds of around 50-70 mph while racing UAVs can reach up to 150 mph. Consequently, even a mere one-second delay translates to a substantial flying distance of 22m to 66m, which poses a significant security risk. Leveraging the computational interdependence of features can contribute to improving detection times, enabling us to achieve rapid and efficient detection capabilities.
Comparing the ablation studies to YOLOv5s, as depicted in Table \ref{t5}, we observe the following trends: i) In Ablation study 1, the MFNet-S model demonstrates a reduction of 33.3\% in preprocessing time and a decrease of 17.5\% in inference time compared to YOLOv5s. ii) For Ablation study 2, the MFNet-S model showcases a preprocessing time reduction of 33.3\%, a remarkable inference time reduction of 60.2\%, and a 33.3\% increase in non-maximum suppression (NMS) operations per image when compared to YOLOv5. iii) In Ablation study 3, the MFNet-FA-S model, in comparison to YOLOv5, achieves a 33.3\% reduction in preprocessing time, a substantial 44.7\% decrease in inference time, and a 33.3\% increase in NMS operations per image. These insights provide a clear understanding of the time-related advantages and efficiencies gained from the ablation studies over the YOLOv5 model.
%\vspace{-3mm}

\begin{table}[!b]
\caption{Time complexity analysis of the proposed models.}
\centering
    \begin{tabular}{p{2cm}p{1.5cm}p{1.5cm}p{1.5cm}}
    \hline
    \textbf{Model} &\textbf{Pre-process (msec)} & \textbf{Inference (msec)} & \textbf{NMS/image (msec)} 
    \\ \hline
       \multicolumn{4}{c}{\textbf{(1) Ablation study 1}}
        % Backbone
        \\ \hline
        \textbf{MFNet-S} &0.3 &8.7  &0.8
        \\ \hline
        \textbf{MFNet-M}&0.3 &9.9  &0.8  
        \\ \hline
        \textbf{MFNet-L}&0.3 &17.7  &0.8
        \\ \hline
        \multicolumn{4}{c}{\textbf{(b) Ablation study 2}}
        \\ \hline
        \textbf{MFNet-S} &0.2 &4.1 & 1.2
        \\ \hline
        \textbf{MFNet-M} &0.4&10.5 &1.3
        \\ \hline
        \textbf{MFNet-L} &0.3 &9.7 &1.7
        \\ \hline
        \multicolumn{4}{c}{\textbf{(b) Ablation study 3}}
        \\ \hline
        \textbf{MFNet-FA-S} &0.2 &5.8 &1.2
        \\ \hline
        \textbf{MFNet-FA-M} &0.2 &6 &1.1
        \\ \hline
        \textbf{MFNet-FA-L} &0.2 &7.3 &1.4
        \\ \hline
        \multicolumn{4}{c}{\textbf{YOLOv5s}}
         \\ \hline
        \textbf{YOLOv5s}  &0.3 &10.3 &0.9
       \\\hline
    \label{t5}
       \end{tabular}
\end{table}

\subsection{Tradeoff in IoU and precision performance}
In Ablation study 1, a noticeable trade-off between IoU and precision performance becomes evident when examining Table \ref{t3} in comparison to the baseline YOLOv5s model. Specifically, the MFNet-S model showcases a decrease of 18.2\% in IoU alongside an increase of 9.6\% in precision relative to YOLOv5s. Moving towards the Ablation study 2, The MFNet-S model reveals an increase of 17.3\% in IoU while experiencing a decrease of 4.6\% in precision compared to YOLOv5s. Finally, for Ablation study 3, the MFNet-FA-L model excels by achieving an impressive 22.6\% increase in IoU due to the FA module. However, it also encounters a marginal reduction of approximately 2.1\% in precision compared to YOLOv5s. These observations offer valuable insights into the performance across different ablation studies in relation to the YOLOv5 baseline model.
\subsection{Comparison with the state-of-the-art schemes}
To ensure a fair comparison, we compared our findings with the most recent literature that utilized the YOLOv5s model for UAV detection. This comparison is then condensed into Table \ref{t6}, presenting an overview of the performance assessment alongside existing state-of-the-art models featured in the latest literature concerning UAV detection.

The proposed models have been assessed across varied datasets with differing input sizes, wherein performance metrics such as precision, recall, mAP, and FPS have been rigorously evaluated. 
Among these models, CT-Net-Middle \cite{lr12} achieves a notable mAP of 95.1\% while operating at a frame rate of 78 FPS. YOLOv5s \cite{lr12}, though slightly lower in mAP at 91.7\%, achieves a faster frame rate of 156 FPS. On the other hand, the Improved YOLOv5s version \cite{s1} attains enhanced precision (93.54\%) and mAP (94.82\%) but omits to report the recall value. Additionally, Fine-tuned YOLOv5x \cite{s2} excels with a precision of 94.7\% and an mAP of 94.1\% on a distinct dataset and input size. SAG-YOLOv5s \cite{s4} stands out with exceptional precision (97.3\%), recall (95.5\%), and mAP (97.6\%) metrics, although it operates at a lower frame rate of 13.2 FPS. In this context, TransVisDrone \cite{s5} achieves a precision of 92\%, recall of 91\%, and an mAP of 95\%, yet does not provide frame rate information. A-YOLOX \cite{s6} excels with a recall of 95.5\% and an mAP of 77.3\%, but precision values remain unreported. Finally, AirDet \cite{s7} achieves a comparatively lower precision of 70.63\%, albeit compensating with a higher frame rate of 1-2 FPS.

When measured against state-of-the-art approaches, the proposed MFNet-S, MFNet-M, and MFNet-L models consistently exhibit competitive precision and mAP performance. While the FPS and parameter counts of the proposed models are relatively higher compared to certain other schemes, this suggests potential computational complexity. The precision values of the proposed models consistently surpass 95\%, underscoring their adeptness in accurately detecting UAVs. Moreover, the mAP values underline the robustness and efficacy of these models in managing intricate detection tasks.

For optimal bird detection performance, the MFNet-FA-M model demonstrates an exceptional precision of 99.8\%, while in UAV detection, the MFNet-L model from the Ablation study stands out with a highly accurate precision of 97.2\%. On the whole, the proposed MFNet-L models from Ablation study 2 present encouraging outcomes, boasting metrics including a precision of 98.4\%, recall of 96.6\%, an mAP of 98.3\%, and an IoU of 72.8\%. In contrast, the MFNet-FA-S model from Ablation study 3 emerges as the main choice, taking into account feature map parameters, computational resources, and efficiency. Additionally, the MFNet-FA-S model stands out as the optimal selection for real-time inference due to its exceptional time efficiency.
 \begin{table}
\caption{Comparison of the proposed models with the state-of-the-art schemes.}
\centering
    \begin{tabular}{p{3cm}p{1.5cm}p{1.5cm}p{1.5cm}}
    \hline
    \textbf{Model} &\textbf{Dataset (images)} & \textbf{Input size} &\textbf{Avg Precision (\%)} 
    \\ \hline
        \textbf{CT-Net-Middle \cite{lr12}}
        &26062 	
        &640$\times$640	
        &95.1	
        \\ \hline
        \textbf{YOLOv5s \cite{lr12}}	
        &26062 	
        &640$\times$640		
        &91.7		
        \\ \hline
        \textbf{Improved YOLOv5s \cite{s1}} 
        &1259 	
        &640$\times$640 	
        &91.09	
        \\ \hline
        \textbf{Fine-tuned YOLOv5x \cite{s2}}	
        &1359 	&416$\times$416 &92.05 
        \\ \hline
        \textbf{YOLOv5s \cite{s3}} 
        &3600	&512$\times$512	&79 
        \\ \hline
        \textbf{SAG-YOLOv5s \cite{s4}}
        &11,286 
        &416$\times$416	 
        &95.5		
        \\ \hline
        \textbf{TransVisDrone \cite{s5}}	
        &2500
        &1280$\times$1280	
        &95		
        \\ \hline
        \textbf{A-YOLOX \cite{s6}}
        &6541 
        &N/G
        &95.5
        \\ \hline
        \textbf{AirDet \cite{s7}}
        &COCO 
        &320$\times$320
        & 70.63 
        \\ \hline
        \textbf{YOLOv5 \cite{aydin2023drone}} &2395 &416$\times$416 &91.8
         \\ \hline
        \textbf{YOLOv3 \cite{dorrer2023solving}} &3548 &1280$\times$720 &97
         \\ \hline 
        \textbf{TransLearn-YOLOx \cite{khan2023translearn}} &11733 &416$\times$416 &94
         \\ \hline           
        \textbf{YOLOv8-M \cite{kim2023high}} &Drone-vs-Bird &640$\times$640 &91.4
        \\ \hline  
        \textbf{TinyFeatureNet \cite{misbah2023tf}} &3000 &416$\times$416 &95.7%
        \\ \hline  
        \textbf{GAANet \cite{10200720}} &5105 &320$\times$320 &96.2
         \\ \hline         
        \multicolumn{4}{c}{\textbf{(1) Ablation study 1}}
        \\\hline
        \textbf{Proposed MFNet-S}	
        &5105 &416$\times$416	&95.2 
        \\ \hline
        \textbf{Proposed MFNet-M}	
        &5105 &416$\times$416	&96.8 
        \\ \hline
        \textbf{Proposed MFNet-L}	
        &5105 &416$\times$416	&95.8 
        \\ \hline
        \multicolumn{4}{c}{\textbf{(2) Ablation study 2}}
        \\ \hline        
        \textbf{Proposed MFNet-S }	
        &5105 &320$\times$320	 &97.1  
        \\ \hline
        \textbf{Proposed MFNet-M }	
        &5105 &320$\times$320	&98.3   
        \\ \hline
        \textbf{Proposed MFNet-L}	
        &5105 &320$\times$320	&\textbf{98.4}  
        \\ \hline    
        \multicolumn{4}{c}{\textbf{(3) Ablation study 3}}
        \\ \hline        
        \textbf{Proposed MFNet-FA-S}	
        &5105 &320$\times$320	&95  
        \\ \hline
        \textbf{Proposed MFNet-FA-M}	
        &5105 &320$\times$320	&96.7
        \\ \hline
        \textbf{Proposed MFNet-FA-L}	
        &5105 &320$\times$320	&97 
        \\ \hline
    \label{t6}
    \vspace{-13mm}
    \end{tabular}
\end{table}
\vspace{-1mm}
\section{Conclusion}
In this paper, we proposed the novel SafeSpace MultifeatureNet (MFNet) architecture that significantly improved the precision and mAP in UAV detection compared to YOLOv5s. To successfully implement the proposed architecture and test its validity in challenging weather conditions, we gathered the existing five datasets of birds and UAVs from the literature to verify its performance on three MFNet/MFNet-FA variants. All algorithms' detection performance was rigorously examined and analyzed with varying environmental backgrounds (i.e., weather conditions) and target scales. Proposed MFNet/MFNet-FA-small, MFNet/MFNet-FA-medium, and MFNet/MFNet-FA-large successfully detected and identified UAVs with the highest UAV detection precision compared to YOLOv5s and the existing state-of-the-art schemes. Thus, the proposed research presents compelling evidence that the proposed MFNet/MFNet-FA models outperform the state-of-the-art schemes for UAV detection. Among the proposed models, MFNet-M from Ablation study 2 achieved an mAP of 97.7\%, the average precision and recall of 98.3\% and 95.4\%, respectively. These results highlight the effectiveness and superiority of the proposed models in UAV detection tasks. Currently, MFNet has the ability to solely recognize birds and drones. Our goal is to expand its capabilities by training multi-class classification and identification models, with the intention of further enhancing their performance.
\bibliographystyle{ieeetr}
\bibliography{MFNet.bib}

\begin{thebibliography}{10}

\bibitem{i1}
Y.~Li, J.~Pawlak, J.~Price, K.~{Al Shamaileh}, Q.~Niyaz, S.~Paheding, and
  V.~Devabhaktuni, ``{Jamming Detection and Classification in OFDM-Based UAVs
  via Feature- and Spectrogram-Tailored Machine Learning},'' {\em IEEE Access},
  vol.~10, pp.~16859--16870, 2022.

\bibitem{i2}
M.~Z. Anwar, Z.~Kaleem, and A.~Jamalipour, ``{Machine Learning Inspired
  Sound-Based Amateur Drone Detection for Public Safety Applications},'' {\em
  IEEE Transactions on Vehicular Technology}, vol.~68, pp.~2526--2534, Mar.
  2019.

\bibitem{i8}
Z.~Kaleem and M.~H. Rehmani, ``{Amateur drone monitoring: State-of-the-art
  architectures, key enabling technologies, and future research directions},''
  {\em IEEE Wireless Communications}, vol.~25, pp.~150--159, Apr. 2018.

\bibitem{i3}
Y.~Zheng, Z.~Chen, D.~Lv, Z.~Li, Z.~Lan, and S.~Zhao, ``{Air-to-air visual
  detection of micro-UAVs: An experimental evaluation of deep learning},'' {\em
  IEEE Robotics and Automation Letters}, vol.~6, pp.~1020--1027, Apr. 2021.

\bibitem{i4}
G.~Lykou, D.~Moustakas, and D.~Gritzalis, ``{Defending Airports from UAS: A
  Survey on Cyber-Attacks and Counter-Drone Sensing Technologies},'' {\em
  Sensors}, vol.~20, pp.~1--40, June 2020.

\bibitem{i5}
J.~McCoy, A.~Rawal, D.~B. Rawat, and B.~M. Sadler, ``{Ensemble Deep Learning
  for Sustainable Multimodal UAV Classification},'' {\em IEEE Transactions on
  Intelligent Transportation Systems}, 2022.

\bibitem{i6}
Y.~Sun, B.~Cao, P.~Zhu, and Q.~Hu, ``{Drone-based RGB-Infrared Cross-Modality
  Vehicle Detection via Uncertainty-Aware Learning},'' {\em IEEE Transactions
  on Circuits and Systems for Video Technology}, 2022.

\bibitem{i7}
J.~V.~V. Gerwen, K.~Geebelen, J.~Wan, W.~Joseph, J.~Hoebeke, and E.~{De
  Poorter}, ``{Indoor Drone Positioning: Accuracy and Cost Trade-Off for Sensor
  Fusion},'' {\em IEEE Transactions on Vehicular Technology}, vol.~71,
  pp.~961--974, Jan. 2022.

\bibitem{yuen18}
H.~Fu, S.~Abeywickrama, L.~Zhang, and C.~Yuen, ``{Low-Complexity Portable
  Passive Drone Surveillance via SDR-Based Signal Processing},'' {\em IEEE
  Communications Magazine}, vol.~56, pp.~112--118, Apr. 2018.

\bibitem{i9}
Y.~Sun, S.~Abeywickrama, L.~Jayasinghe, C.~Yuen, J.~Chen, and M.~Zhang,
  ``{Micro-Doppler Signature-Based Detection, Classification, and Localization
  of Small UAV with Long Short-Term Memory Neural Network},'' {\em IEEE
  Transactions on Geoscience and Remote Sensing}, vol.~59, pp.~6285--6300, Aug.
  2021.

\bibitem{lr3}
R.~Kılı{\c{c}}, N.~Kumbasar, E.~A. Oral, and I.~Y. Ozbek, ``{Drone
  classification using RF signal based spectral features},'' {\em Engineering
  Science and Technology, an International Journal}, vol.~28, p.~101028, Apr.
  2022.

\bibitem{xie2020adaptive}
J.~Xie, J.~Yu, J.~Wu, Z.~Shi, and J.~Chen, ``Adaptive switching
  spatial-temporal fusion detection for remote flying drones,'' {\em IEEE
  Transactions on Vehicular Technology}, vol.~69, no.~7, pp.~6964--6976, 2020.

\bibitem{lr4}
H.~R. Alsanad, A.~Z. Sadik, O.~N. Ucan, M.~Ilyas, and O.~Bayat, ``{YOLO-V3
  based real-time drone detection algorithm},'' {\em Multimedia Tools and
  Applications}, vol.~81, pp.~26185--26198, July 2022.

\bibitem{lr1}
S.~Dogru and L.~Marques, ``{Drone Detection Using Sparse LIDAR Measurements},''
  {\em IEEE Robotics and Automation Letters}, vol.~7, pp.~3062--3069, Apr.
  2022.

\bibitem{lr2}
O.~O. Medaiyese, M.~Ezuma, A.~P. Lauf, and I.~Guvenc, ``{Wavelet transform
  analytics for RF-based UAV detection and identification system using machine
  learning},'' {\em Pervasive and Mobile Computing}, vol.~82, p.~101569, June
  2022.

\bibitem{lr5}
M.~Wisniewski, Z.~A. Rana, and I.~Petrunin, ``{Drone Model Classification Using
  Convolutional Neural Network Trained on Synthetic Data},'' {\em Journal of
  Imaging}, vol.~8, p.~218, Aug. 2022.

\bibitem{lr7}
X.~Dai and M.~Nagahara, ``{Platooning control of drones with real-time deep
  learning object detection},'' {\em Advanced Robotics}, 2022.

\bibitem{lr8}
Q.~Dong, Y.~Liu, and X.~Liu, ``{Drone sound detection system based on feature
  result-level fusion using deep learning},'' {\em Multimedia Tools and
  Applications}, pp.~1--23, June 2022.

\bibitem{lr9}
H.~C. Kumawat, M.~Chakraborty, and A.~{Arockia Bazil Raj}, ``{DIAT-RadSATNet-A
  Novel Lightweight DCNN Architecture for Micro-Doppler-Based Small Unmanned
  Aerial Vehicle (SUAV) Targets' Detection and Classification},'' {\em IEEE
  Transactions on Instrumentation and Measurement}, vol.~71, 2022.

\bibitem{lr10}
M.~Elsayed, A.~S. Mashaly, M.~Reda, and A.~S. Amein, ``{Visual Drone Detection
  In Static Complex Environment},'' {\em 13th International Conference on
  Electrical Engineering (ICEENG)}, pp.~154--158, 2022.

\bibitem{lr11}
T.~Ye, W.~Qin, Y.~Li, S.~Wang, J.~Zhang, and Z.~Zhao, ``{Dense and Small Object
  Detection in UAV-Vision Based on a Global-Local Feature Enhanced Network},''
  {\em IEEE Transactions on Instrumentation and Measurement}, vol.~71, 2022.

\bibitem{lr12}
T.~Ye, J.~Zhang, Y.~Li, X.~Zhang, Z.~Zhao, and Z.~Li, ``{CT-Net: An Efficient
  Network for Low-Altitude Object Detection Based on Convolution and
  Transformer},'' {\em IEEE Transactions on Instrumentation and Measurement},
  vol.~71, 2022.

\bibitem{f1}
C.~Y. Wang, H.~Y. {Mark Liao}, Y.~H. Wu, P.~Y. Chen, J.~W. Hsieh, and I.~H.
  Yeh, ``{CSPNet: A New Backbone that can Enhance Learning Capability of
  CNN},'' {\em IEEE Computer Society Conference on Computer Vision and Pattern
  Recognition Workshops}, vol.~2020, pp.~1571--1580, Nov. 2019.

\bibitem{f2}
K.~He, X.~Zhang, S.~Ren, and J.~Sun, ``{Spatial Pyramid Pooling in Deep
  Convolutional Networks for Visual Recognition},'' {\em Lecture Notes in
  Computer Science (including subseries Lecture Notes in Artificial
  Intelligence and Lecture Notes in Bioinformatics)}, vol.~8691 LNCS,
  pp.~346--361, June 2014.

\bibitem{f3}
T.-Y. Lin, P.~Doll{\'{a}}r, R.~Girshick, K.~He, B.~Hariharan, and S.~Belongie,
  ``{Feature Pyramid Networks for Object Detection},'' Dec. 2016.

\bibitem{f4}
S.~Liu, L.~Qi, H.~Qin, J.~Shi, and J.~Jia, ``{Path Aggregation Network for
  Instance Segmentation},'' {\em Proceedings of the IEEE Computer Society
  Conference on Computer Vision and Pattern Recognition}, pp.~8759--8768, Mar.
  2018.

\bibitem{f6}
I.~S. Isa, M.~S.~A. Rosli, U.~K. Yusof, M.~I.~F. Maruzuki, and S.~N. Sulaiman,
  ``{Optimizing the Hyperparameter Tuning of YOLOv5 for Underwater
  Detection},'' {\em IEEE Access}, vol.~10, pp.~52818--52831, 2022.

\bibitem{10189370}
Y.~Zheng, Y.~Zhan, X.~Huang, and G.~Ji, ``Yolov5s fmg: An improved small target
  detection algorithm based on yolov5 in low visibility,'' {\em IEEE Access},
  vol.~11, pp.~75782--75793, 2023.

\bibitem{s1}
B.~Liu and H.~Luo, ``{An Improved YOLOv5 for Multi-Rotor UAV Detection},'' {\em
  Electronics}, vol.~11, p.~2330, Jul. 2022.

\bibitem{s2}
N.~Al-Qubaydhi, A.~Alenezi, T.~Alanazi, A.~Senyor, N.~Alanezi, B.~Alotaibi,
  M.~Alotaibi, A.~Razaque, A.~A. Abdelhamid, and A.~Alotaibi, ``{Detection of
  Unauthorized Unmanned Aerial Vehicles Using YOLOv5 and Transfer Learning},''
  {\em Electronics}, vol.~11, p.~2669, Aug. 2022.

\bibitem{s4}
D.~Zarpalas, A.~Dimou, A.~Schumann, L.~Sommer, A.~Fascista, Y.~Lv, Z.~Ai,
  M.~Chen, X.~Gong, Y.~Wang, and Z.~Lu, ``{High-Resolution Drone Detection
  Based on Background Difference and SAG-YOLOv5s},'' {\em Sensors}, vol.~22,
  p.~5825, Aug. 2022.

\bibitem{s5}
T.~Sangam, I.~R. Dave, W.~Sultani, and M.~Shah, ``{TransVisDrone:
  Spatio-Temporal Transformer for Vision-based Drone-to-Drone Detection in
  Aerial Videos},'' {\em arXiv preprint arXiv:2210.08423}, 2022.

\bibitem{s6}
Y.~Xu, D.~Zhong, J.~Zhou, Z.~Jiang, Y.~Zhai, and Z.~Ying, ``A novel uav visual
  positioning algorithm based on a-yolox,'' {\em Drones}, vol.~6, no.~11,
  p.~362, 2022.

\bibitem{s7}
B.~Li, C.~Wang, P.~Reddy, S.~Kim, and S.~Scherer, ``Airdet: Few-shot detection
  without fine-tuning for autonomous exploration,'' in {\em Computer
  Vision--ECCV 2022: 17th European Conference, Tel Aviv, Israel, October
  23--27, 2022, Proceedings, Part XXXIX}, pp.~427--444, Springer, 2022.

\bibitem{s3}
V.~T. Hai, V.~N. Le, D.~Q. Khanh, N.~P. Nam, and D.~V. Sang, ``{Multi-size
  drone detection using YOLOv5 network},'' {\em Journal of Military Science and
  Technology}, pp.~142--148, June 2022.

\bibitem{aydin2023drone}
B.~Aydin and S.~Singha, ``Drone detection using yolov5,'' {\em Eng}, vol.~4,
  no.~1, pp.~416--433, 2023.

\bibitem{dorrer2023solving}
M.~G. Dorrer and A.~E. Alekhina, ``Solving the problem of biodiversity analysis
  of bird detection and classification in the video stream of camera traps,''
  in {\em E3S Web of Conferences}, vol.~390, EDP Sciences, 2023.

\bibitem{khan2023translearn}
M.~U. Khan, M.~Dil, M.~Misbah, F.~A. Orakazi, M.~Z. Alam, and Z.~Kaleem,
  ``Translearn-yolox: Improved-yolo with transfer learning for fast and
  accurate multiclass uav detection,'' in {\em 2023 International Conference on
  Communication, Computing and Digital Systems (C-CODE)}, pp.~1--7, IEEE, 2023.

\bibitem{kim2023high}
J.-H. Kim, N.~Kim, and C.~S. Won, ``High-speed drone detection based on
  yolo-v8,'' in {\em ICASSP 2023-2023 IEEE International Conference on
  Acoustics, Speech and Signal Processing (ICASSP)}, pp.~1--2, IEEE, 2023.

\bibitem{misbah2023tf}
M.~Misbah, M.~U. Khan, Z.~Yang, and Z.~Kaleem, ``Tf-net: Deep learning
  empowered tiny feature network for night-time uav detection,'' in {\em
  International Conference on Wireless and Satellite Systems}, pp.~3--18,
  Springer, 2023.

\bibitem{10200720}
M.~U. Khan, M.~Misbah, Z.~Kaleem, Y.~Deng, and A.~Jamalipour, ``Gaanet: Ghost
  auto anchor network for detecting varying size drones in dark,'' in {\em 2023
  IEEE 97th Vehicular Technology Conference (VTC2023-Spring)}, pp.~1--5, 2023.

\end{thebibliography}
\begin{IEEEbiography}[{\includegraphics[width=1in,height=2.75in,clip,keepaspectratio]{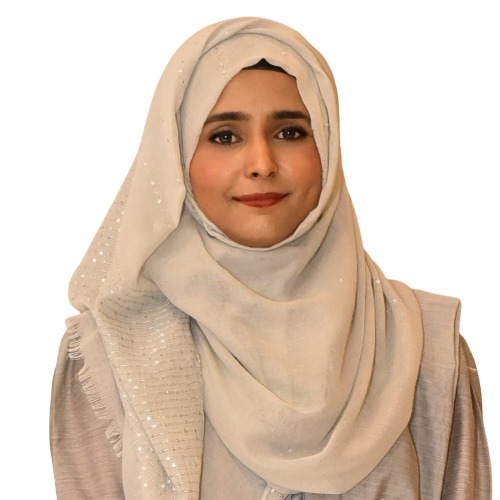}}]{Misha Urooj Khan} received her master's degree in electronics engineering with a specialization in electronic system design from UET Taxila. She is the Chairperson of the Community of Research and Development (CRD). Her academic pursuits have resulted in 30 publications to date, focusing on various topics, including bio-medical engineering, machine learning, deep learning, audio processing, computer vision, predictive analytics and quantum machine learning.
\end{IEEEbiography}
\vspace{-13mm}
\begin{IEEEbiography}[{\includegraphics[width=1in,height=1.25in,clip,keepaspectratio]{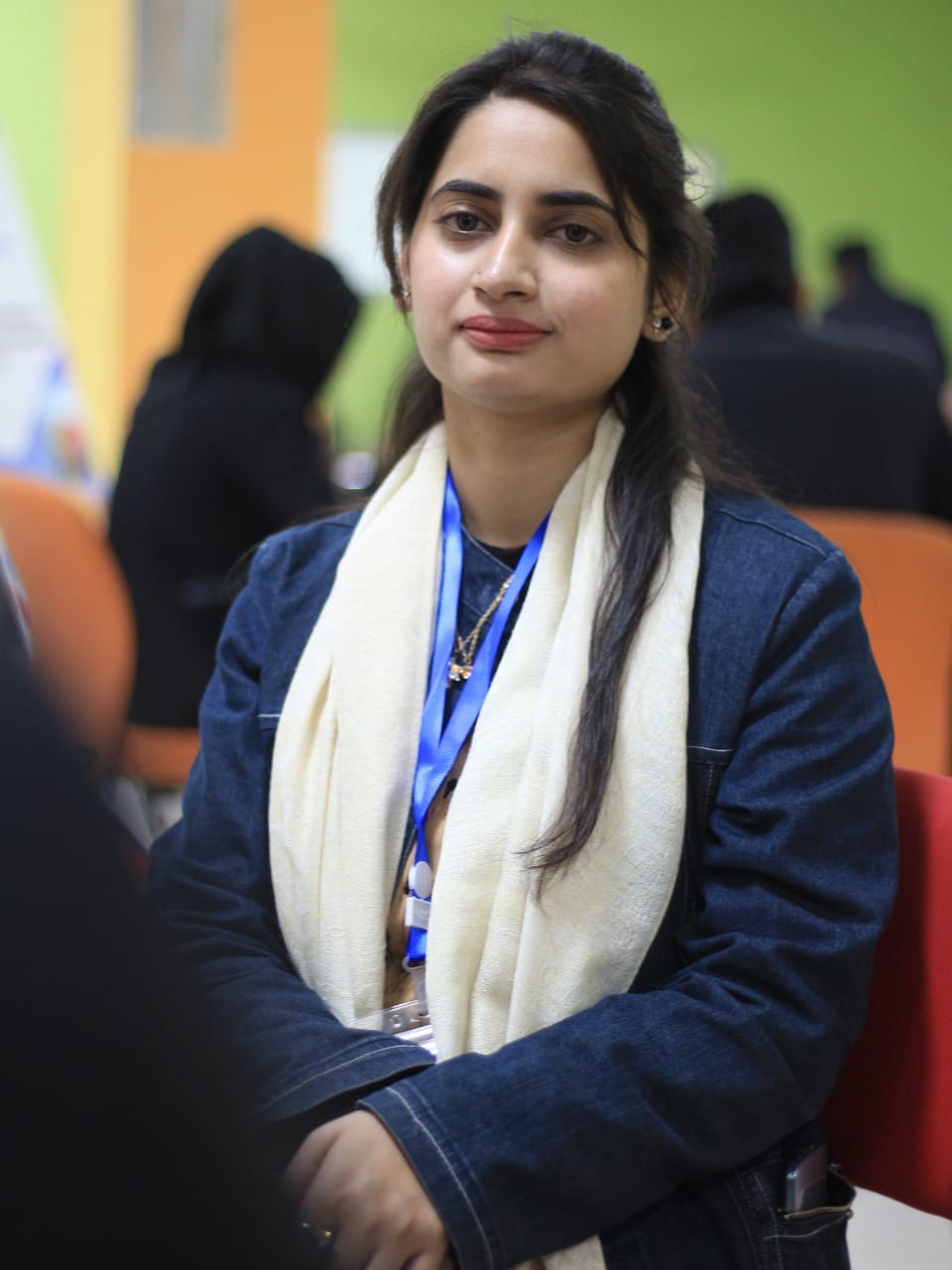}}]{Mahnoor Dil}
received the BS and MS  degree in Electrical Engineering with specialization in Computer from COMSATS University Islamabad, Wah Campus Pakistan. She is currently working as a Research Assistant at COMSATS University Islamabad, Wah Campus under HEC funded project to develop UAV detection system using computer vision models.
\end{IEEEbiography}
\vspace{-13mm}
\begin{IEEEbiography}[{\includegraphics[width=1in,height=1.25in,clip,keepaspectratio]{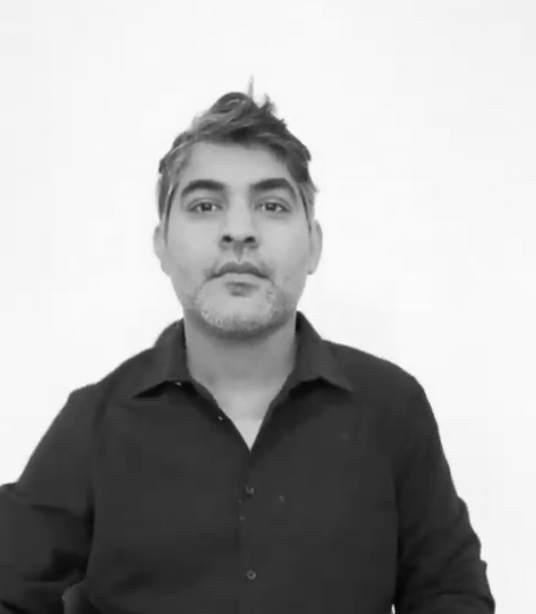}}]{Muhamad Zeshan Alam} received  his  B.S.  degree  in Computer  Engineering from  COMSATS  University,  Pakistan,   M.S.  degree  in  Electrical  and Electronics  Engineering  from  the  University  of Bradford, UK, and Ph.D. In Electrical Engineering  and  Cyber-Systems  from  Istanbul  Medipol University, Turkey. He worked at the University of Cambridge as a postdoctoral fellow where his work focused on computer vision and machine learning models. He recently joined Brandon University, Canada, as an assistant professor while also working as a Computer Vision Consultant at Vimmerse INC. His research interests include immersive videos, computational imaging, computer vision and machine learning modeling.
\end{IEEEbiography}
\vspace{-11mm}
\begin{IEEEbiography}[{\includegraphics[width=1in,height=1.25in,clip,keepaspectratio]{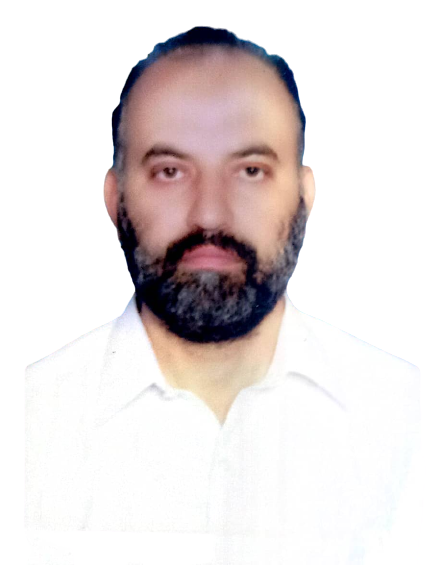}}]{Farooq Alam Orakzai} is an Assistant Professor in the Electrical and Computer Engineering Department, COMSATS University Islamabad, Wah Campus. He published several research articles. His current research interest includes 5G technologies, unmanned aerial vehicles (UAVs), computer vision, cognitive radio networks, wireless mobile communication, wireless sensor networks, and optical fiber communication.
\end{IEEEbiography}
\vspace{-15mm}
\begin{IEEEbiography}[{\includegraphics[width=1in,height=1.25in,clip,keepaspectratio]{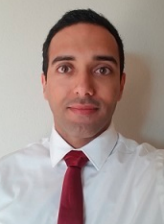}}]{Abdullah M. Almasoud} received the B.Sc. degree in computer engineering from King Saud University, Riyadh, Saudi Arabia, and the M.Sc. degree in computer engineering and the Ph.D. degree in computer engineering and electrical engineering from Iowa State University, Ames, IA, USA. Currently, he is an Assistant Professor with the Department of Electrical Engineering, Prince Sattam Bin Abdulaziz University, Al-Kharj, Saudi Arabia. His research interests include wireless networks, cognitive radio networks, the Internet of Things, UAV-assisted networking, and RF energy harvesting. He was a recipient of the Best Paper Award at the IEEE Global Communications Conference 2018 on Ad Hoc and Sensors Networks Symposium.
\end{IEEEbiography}
\begin{IEEEbiography}[{\includegraphics[width=1in,height=1.25in,clip,keepaspectratio]{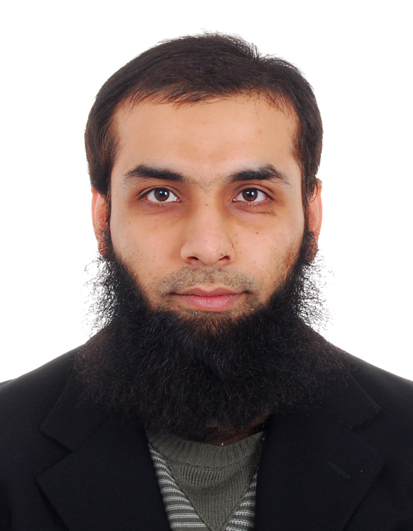}}]{Zeeshan Kaleem [Senior Member, IEEE]} is serving as an Associate Professor at COMSATS University Islamabad, Wah Campus. He consecutively received the National Research Productivity Award (RPA) awards from the Pakistan Council of Science and Technology (PSCT) in 2017 and 2018. He won the Higher Education Commission (HEC) Best Innovator Award for the year 2017, and there was a single award all over Pakistan. He is the recipient of the 2021 Top Reviewer Recognition Award for IEEE TRANSACTIONS on VEHICULAR TECHNOLOGY and published over 100 technical papers and patents.
\end{IEEEbiography}
\vspace{-11mm}
\begin{IEEEbiography}[{\includegraphics[width=1in,height=1.25in,clip,keepaspectratio]{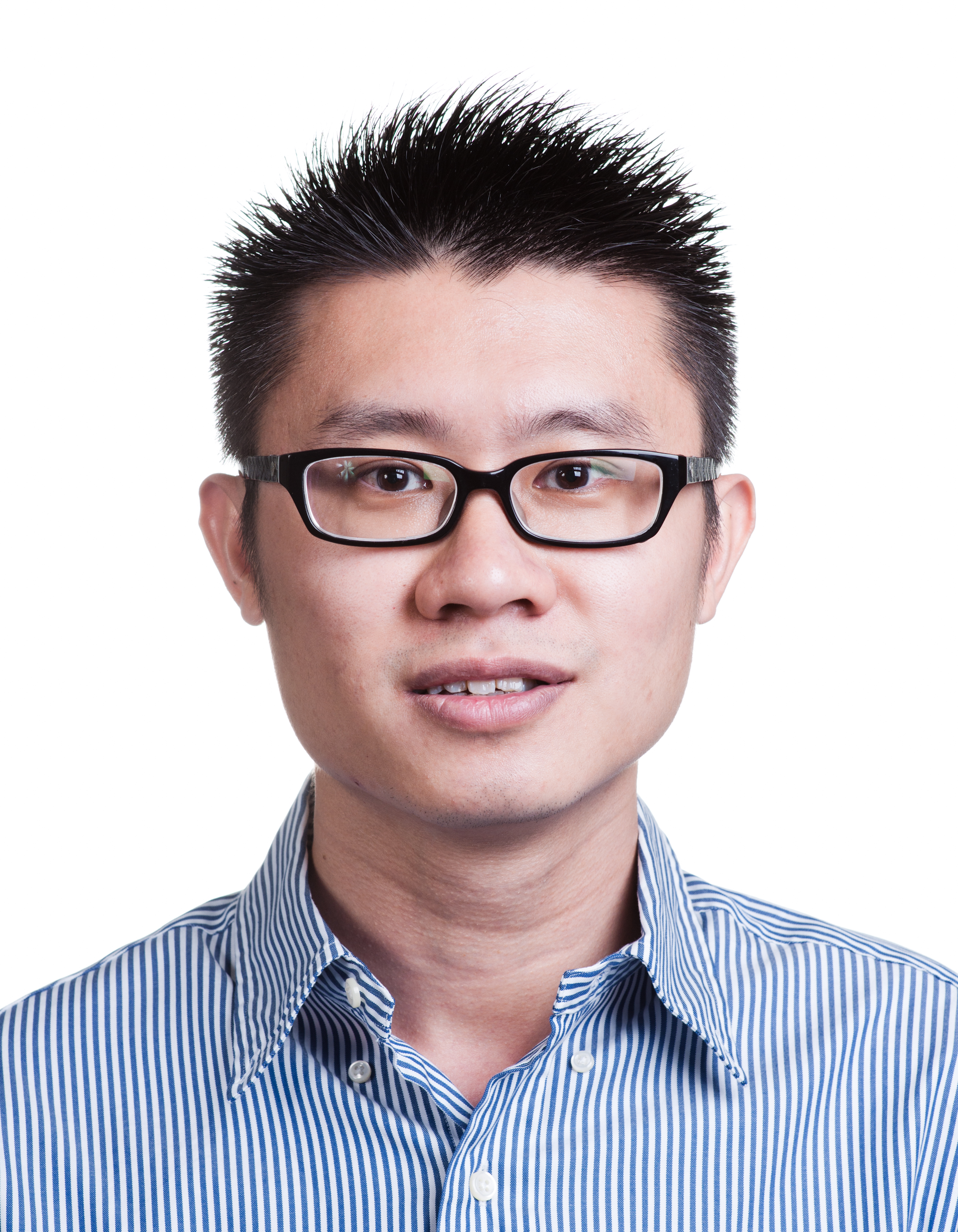}}]{Chau Yuen (S’02-M’06-SM’12-F’21)} received the B.Eng. and Ph.D. degrees from Nanyang Technological University (NTU), Singapore, in 2000 and 2004, respectively. He received the IEEE Asia Pacific Outstanding Young Researcher Award in 2012 and IEEE VTS Singapore Chapter Outstanding Service Award on 2019. Currently, he serves as an Editor for the IEEE TRANSACTIONS ON VEHICULAR TECHNOLOGY, IEEE System Journal, and IEEE Transactions on Network Science and Engineering. He is a Distinguished Lecturer of IEEE Vehicular Technology Society.
\end{IEEEbiography}

\end{document}